\documentclass[journal]{IEEEtran}
\usepackage{graphicx}
\usepackage{multirow}
\usepackage{epstopdf}
\usepackage{times}
\usepackage{amsmath}
\usepackage{amssymb}
\usepackage{amsfonts}
\usepackage{color}
\usepackage{subfigure}
\usepackage{hyperref}

\usepackage{algorithm}
\usepackage{algorithmic}
\usepackage{rotating}
\usepackage[export]{adjustbox} 

\usepackage{adjustbox,lipsum}
\usepackage{dsfont}
\usepackage{booktabs}
\usepackage{multirow}
\usepackage{array}
\usepackage{color}

\newcommand{\etal}{{\em et al.\,}}       
\newcommand{\eg}{{\em e.g.}}           
\newcommand{\ie}{{\em i.e.}}           
\newcommand{\etc}{{\em etc}}

\begin{document}

\title{Feature-based Style Randomization for Domain Generalization}

\author{Yue Wang$^\dagger$,
        Lei Qi$^\dagger$,
        Yinghuan Shi$^\star$,
        Yang Gao
\thanks{The work was supported by National Key Research and Development Program of China (2019YFC0118300), China Postdoctoral Science Foundation Project (2021M690609), Jiangsu Natural Science Foundation Project (BK20210224) and CAAI-Huawei MindSpore Project (CAAIXSJLJJ-2021-042A).}
\thanks{Yue Wang, Yinghuan Shi and Yang Gao are with the National Key Laboratory for Novel Software Technology and the National Institute of Healthcare Data Science, Nanjing University, Nanjing, China, 210023 (e-mail: wyue@smail.nju.edu.cn; syh@nju.edu.cn; gaoy@nju.edu.cn).}
\thanks{Lei Qi is with the School of Computer Science and Engineering, and the Key Lab of Computer Network and Information Integration (Ministry of Education), Southeast University, Nanjing, China, 211189 (e-mail: qilei@seu.edu.cn).}
\thanks{Yue Wang and Lei Qi are the co-first author.}
\thanks{Corresponding author: Yinghuan Shi.}


}

%
%

\markboth{~}%
{Shell \MakeLowercase{\textit{et al.}}: Bare Demo of IEEEtran.cls for IEEE Journals}

\maketitle

\begin{abstract}
		As a recent noticeable topic, domain generalization (DG) aims to first learn a generic model on multiple source domains and then directly generalize to an arbitrary unseen target domain without any additional adaption. In previous DG models, by generating virtual data to supplement observed source domains, the data augmentation based methods have shown its effectiveness. To simulate the possible unseen domains, most of them enrich the diversity of original data via image-level style transformation. However, we argue that the potential styles are hard to be exhaustively illustrated and fully augmented due to the limited referred styles, leading the diversity could not be always guaranteed. Unlike image-level augmentation, we in this paper develop a simple yet effective feature-based style randomization module to achieve feature-level augmentation, which can produce random styles via integrating random noise into the original style. Compared with existing image-level augmentation, our feature-level augmentation favors a more goal-oriented and sample-diverse way. Furthermore, to sufficiently explore the efficacy of the proposed module, we design a novel progressive training strategy to enable all parameters of the network to be fully trained. Extensive experiments on three standard benchmark datasets, \ie, PACS, VLCS and Office-Home, highlight the superiority of our method compared to the state-of-the-art methods.
\end{abstract}


%
\IEEEpeerreviewmaketitle

\section{Introduction}
\IEEEPARstart{M}{ost} of machine learning algorithms are often incapable of handling the domain-shift case, where training data and test data are from the different distributions~\cite{Li2017DeeperBA,Balaji2018MetaRegTD,Patel2015VisualDA, Taylor2009TransferLF,9530705,9555613,9514559}. 
	To address this issue, unsupervised domain adaptation (UDA) is has been proposed to alleviate the data-distribution discrepancy by utilizing labeled source domain and unlabeled target domain to jointly train the model~\cite{Long2014TransferJM,Long2015LearningTF,Ganin2015UnsupervisedDA,Daum2010CoregularizationBS,Ganin2016DomainAdversarialTO,Hoffman2018CyCADACA, Long2016UnsupervisedDA, Luo2019TakingAC, Saenko2010AdaptingVC}. 
	Thanks to these efforts, the performance of learning under distribution discrepancy has now been greatly improved. Despite their success, directly deploying a trained UDA model to a new scenario (\ie, target domain) is inconvenient and sometimes even inapplicable. Firstly, the trained UDA model should be re-trained by incorporating the unlabeled data in new target domains. Moreover, in some scenarios, \eg, 1) massive samples in new target domain or 2) target domain could not be fully accessed, the required re-training process seems hard to realize. Being aware of this fact, domain generalization (DG)---the model trained on multiple source domains is expected to generalize well to unseen target domains without any adaptation~\cite{Blanchard2011GeneralizingFS,Muandet2013DomainGV}---has been received considerable attention in recent years. The ultimate goal of DG is to avoid recollecting and re-training data for new scenarios, which is usually realized by enhancing the generalization ability of the trained model in previously unseen domains~\cite{Balaji2018MetaRegTD,Li2018DomainGW,Dou2019DomainGV,Li2019FeatureCriticNF,Zhao2020LearningTG,Rahman2020CorrelationawareAD}.

	\begin{figure}
		\begin{center}
			\subfigure[Image-based augmentation method.]{
				\label{intro.sub.1}	\includegraphics[width=8.0cm]{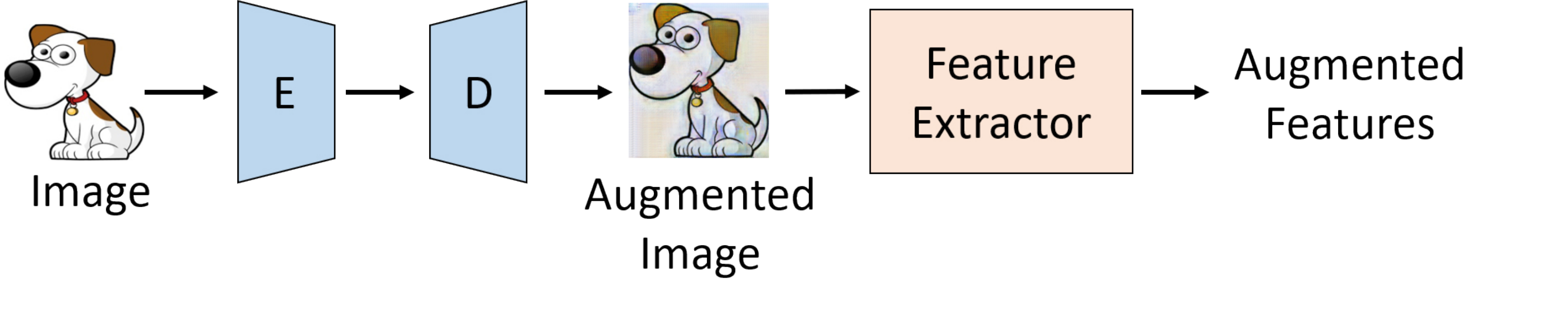}}\vspace{-2mm}
			\subfigure[Feature-based augmentation method.]{
				\label{intro.sub.2}	\includegraphics[width=8.0cm]{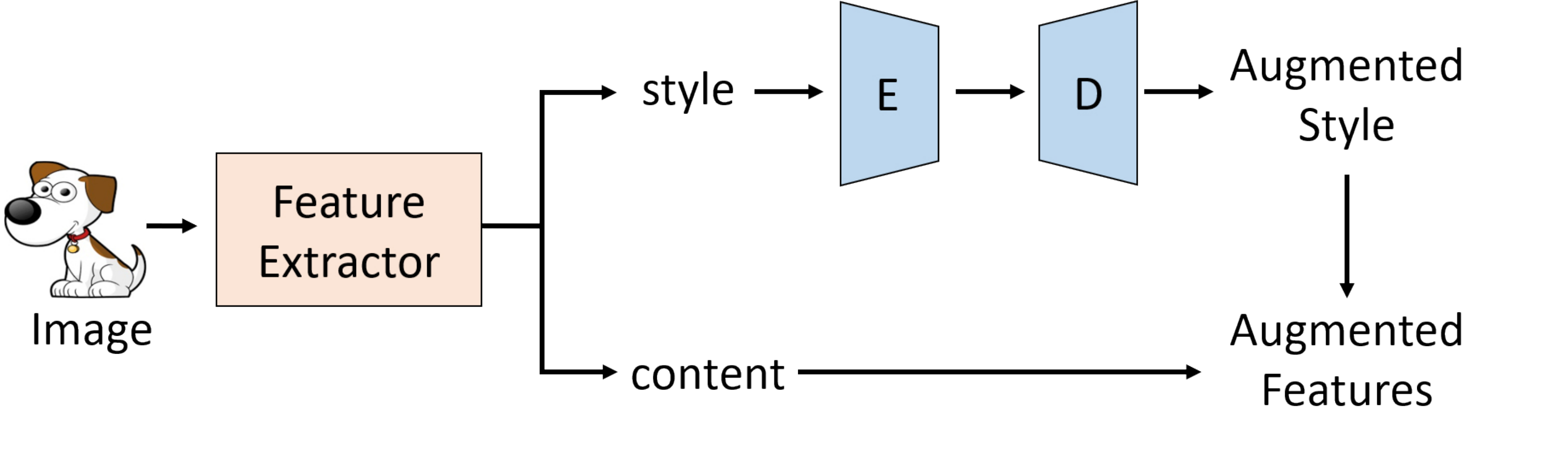}}
			\caption{The comparison between our feature-based augmentation method and the image-based augmentation method for DG. Image-based augmentation method generate different views of an input image via encoder-decoder network and then obtain the augmented features for training according to the feature extractor, which are limited to operations in the image space. We propose an additional feature-based augmentation that directly operates in the abstract feature space, which is a more goal-oriented and  sample-diverse way.}
			\label{Fig:intro}
		\end{center}
	\end{figure}

	Among these recent DG models, the data augmentation based DG methods have shown their promising performance~\cite{Shankar2018GeneralizingAD,Somavarapu2020FrustratinglySD,Zhou2020DeepDI,Zhou2020LearningTG} by employing learning-based augmentation strategy to generate diverse data. The generated data can be regarded as a complement to original data by simulating the data from possible unseen domains to aid the learning process. 
	However, most existing data augmentation based DG methods only focus on the image-level augmentation in source domains by directly generating virtual images, \ie, translating images based on the given styles collected from the auxiliary dataset~\cite{Yue2019DomainRA} or converting images into other styles that are different from the styles in the training domains~\cite{Zhou2020DeepDI,Zhou2020LearningTG}.
	Unfortunately, previous image-level augmentation adopts limited styles, 
	while all the real styles in unseen target domains are hard to be exhaustively illustrated and fully augmented. In this meaning, the sample diversity could not be always guaranteed, which might occur performance degeneration especially there exists large difference between the generated virtual style and real unseen style. Differently, our goal is to enhance the diversity of images from the feature augmentation view. For illustration, the comparison between image-based and our feature-based augmentation methods is shown in Fig.~\ref{Fig:intro}. 

	
	
	Besides, most image-level augmentation based methods 
	utilize the GANs-based model in the training stage, which is not readily to be trained~\cite{Yue2019DomainRA}. 
	To alleviate the aforementioned issues, considering that the statistics from convolutional neural networks (CNNs) could indicate the representative style information~\cite{Huang2017ArbitraryST}, some works like MixStyle~\cite{zhou2021domain} are proposed to enrich the style diversity from the feature augmentation perspective. Although MixStyle does not employ the generative model in the training phase, it produces the new style via merely mixing the existing styles from few observed source domains, leading the insufficient diversity still occurs.

	To be specific, we propose a novel Feature-based Style Randomization (abbreviated as FSR) module for domain generalization, where the feature of an original sample can be converted into a new feature with a random style. To achieve this goal, we employ the encoder-decoder network to integrate the noise information into the original style information in a hidden space. Since the proposed module is learned and manipulated in the feature space by introducing additional noise information, it can perform more random and diverse transformations for the input images. Compared with conventional augmentation based methods, our proposed way is more goal-oriented for DG problem. 
	Moreover, 
	we further develop a novel progressive training scheme for the proposed module to ensure that all parts of the model can be fully trained. Extensive experiments on three public available datasets validate the effectiveness of both the proposed module and the progressive training scheme.
	
	Our contributions are summarized as below:
    
	\begin{itemize}
		\item We propose a novel module for DG task, namely Feature-based Style Randomization (FSR), which can produce random styles to enhance the model robustness and effectiveness in unseen domains. 
		\item 
		We design a novel and effective progressive training strategy to better leverage our proposed module during the training course, which can guarantee that all parts of the network are sufficiently trained.
		\item We evaluate our approach on three standard benchmark datasets, \ie, PACS, OfficeHome and VLCS. The experimental results show that our approach achieves the state of the art accuracy on all datasets. 
	\end{itemize}

	The rest of this paper is organized as follows. In Section II, we review the related work about unsupervised domain adaptation, domain generalization, domain randomization and style transfer. The technical details in our proposed method, \eg, feature based style randomization, and progressive training scheme, are introduced in Section III. In Section IV, we report the experimental results, ablation study and further analysis for our proposed method. Finally, we conclude this paper in Section V.


\section{Related Work}
In this section, we review the related work about unsupervised domain adaptation, domain generalization, domain randomization and style transfer.

\subsection{Unsupervised Domain Adaptation}
	Unsupervised Domain Adaptation (UDA) aims to tackle the problem of domain shift between the source and target data by learning domain invariant/aligned features from the labeled source domain and the unlabelled target domain, which is closely relevant to domain generalization. The mainstream methods usually align the distributions between source and target domains in the input space~\cite{Hoffman2018CyCADACA,Sankaranarayanan2018LearningFS,Li2019BidirectionalLF}, feature space~\cite{Daum2010CoregularizationBS,Ganin2016DomainAdversarialTO,Long2016UnsupervisedDA,Saenko2010AdaptingVC,Tzeng2017AdversarialDD,9495801}, or output space~\cite{Luo2019TakingAC,Saito2018MaximumCD} by minimize Maximum Mean Discrepancy (MMD) or adopting adversarial learning. Li \etal \cite{Li2019BidirectionalLF} propose a novel bidirectional image-to-image translation learning framework for domain adaptation of segmentation. Li \etal \cite{9495801} introduce a multi-view imaginative reasoning network to encourage the encoder to obtain the strong multi-view imaginative reasoning ability and get domain-invariant features by triple adversarial learning.
    In addition, some methods employ self-training to iteratively increase a set of labeled instances by generating pseudo-labels for unlabeled target instances~\cite{Zou2018UnsupervisedDA,Saleh2018EffectiveUO,Zhong2019InvarianceME,Zou2019ConfidenceRS,Guan2021ScaleVM,Guan2019UnsupervisedDA,9559925}. Zou \etal \cite{Zou2018UnsupervisedDA} propose to generate pseudo labels on target data which is constrained by class-balanced framework and  spatial priors and re-training the model with these labels. The difference between UDA and DG is that UDA has to access to the target domain while DG cannot observe any target domain during training. This makes DG more challenging, but more realistic and favorable than DA in practical applications. 

	\subsection{Domain Generalization}
	Existing works for DG can be summarized into three categories: domain-invariant-representation-learning-based methods, meta-learning-based methods and data-augmentation-based methods.
	
	\textbf{Domain-invariant representation learning.} Inspired by unsupervised domain adaptation methods~\cite{Long2014TransferJM,Long2015LearningTF,Ganin2015UnsupervisedDA, Ganin2016DomainAdversarialTO}, some DG methods resort to learn the domain-invariant feature to mitigate the domain gap. Muandet \etal \cite{Muandet2013DomainGV} propose a kernel-based optimization algorithm that learns an invariant transformation by minimizing the dissimilarity across domains. Li \etal \cite{Li2018DeepDG} consider the conditional distribution and minimize discrepancy of a joint distribution to get the general feature representation. Also, \cite{Li2018DomainGW} optimizes a multi-domain autoencoder regularized by the MMD distance, a discriminator and a classifier in an adversarial training manner. Moreover, Seo \etal \cite{Seo2019LearningTO} develop to learn domain-specific optimized normalization layers which identify the best combination of batch normalization (BN) and instance normalization (IN) because IN could remove the domain-specific information effectively. Nam \etal \cite{nam2021reducing} propose  Style-Agnostic Network to disentangle style encodings from class categories to make focus more on the contents by adversarial learning.  
	
	\textbf{Meta learning.} Recently, meta-learning has been used to address DG by splitting the source domains into meta-train and meta-test to simulate the domain shift during training. Li \etal \cite{Li2018LearningTG} extend MAML \cite{Finn2017ModelAgnosticMF} to optimize the model on the meta-train to improve the performance of the meta-test. Balaji \etal \cite{Balaji2018MetaRegTD} capture the notion of domain generalization through a regularizer learned based on the meta-learning framework. Besides, Li \etal \cite{Li2019FeatureCriticNF} introduce a feature-critic network to train a domain-invariant feature extractor by a meta-learning regularizer. And, Bai \etal \cite{bai2021ood} develop a neural architecture search method which optimizes the architecture with respect to its performance on generated data by gradient descent.

	\textbf{Data augmentation.} Our work is most relevant to the data augmentation based methods. Recent data augmentation based methods enrich the diversity of training data by utilizing image generation or transformation of the input data. Shankar \etal \cite{Shankar2018GeneralizingAD} introduce CrossGrad to perturb input instances based on domain label, which is inspired by the adversarial attacks \cite{Goodfellow2015ExplainingAH}. However, the perturbations are simply depended on the gradient as adversarial attacks, so that the transformation of the input images is imperceptible and lack of texture changes to simulate the domain diversity.
	Recently, Zhou \etal \cite{Zhou2020DeepDI} develop deep domain-adversarial image generation (DDAIG) to map the source training data to unseen domains by a learned transformation network. Besides, the method in \cite{Zhou2020LearningTG} learns a conditional generator network using optimal transport (OT)-based distribution divergence to synthesize unseen domain images. Most of these methods are based on image-to-image transformation, thus the operation is complicated and the variation of the style information from source domains is limited \cite{zhou2021domain, li2021simple}. 
	Besides, there are some works concerned on feature augmentations through adversarial perturbation \cite{Qiao2021UncertaintyguidedMG, Liu2021DomainGV}. 
	
	However, these methods aim to produce images with different styles from the source domains and use both the original images and the augmented images to train the feature extractor and classifier as shown in Fig. \ref{Fig:intro} or create feature augmentations by directly adding perturbation to the original features. In this way, they not only augment the input but also yield label uncertainty. Our method disentangles the features into two parts, \ie, style and content, with only perturbing the style information of original domains with random noise to synthesize novel features, which is a more goal-oriented way. Moreover, instead of generating limited novel styles according to the given source domains \cite{Shankar2018GeneralizingAD, Zhou2020DeepDI, Zhou2020LearningTG}, our method learns to generate random, novel and domain-specific styles by introducing extra random noise in the hidden space generated by employing an auto-encoder module, which is a much more sample-diverse way and can comprehensively cover more domain variations for DG task.

\subsection{Domain Randomization}
Our approach is inspired by domain randomization (DR), which has been recently successfully used in semantic segmentation \cite{Zakharov2019DeceptionNetND}, object detection \cite{Prakash2019StructuredDR} and 6D pose estimation \cite{Sundermeyer2018Implicit3O,Tremblay2018TrainingDN}. Tobin \textit{et al.} \cite{Tobin2017DomainRF} explore domain randomization, which generates simulated images by rendering background, changing illumination, transforming colorization and trains models on them to reduce the gap between simulated images and real images. Instead of placing objects and distractors randomly, Prakash \textit{et al.} \cite{Prakash2019StructuredDR} propose the structured domain randomization, which randomly places objects and distractors according to the probability distribution generated by the specific problem at hand. Moreover, Zakharov \textit{et al.} \cite{Zakharov2019DeceptionNetND} design a min-max optimization scheme to guide the generation of synthetic renderings. Different from the above general DR-based methods, which are based on programmatic simulators~\cite{Zakharov2019DeceptionNetND,Prakash2019StructuredDR,Sundermeyer2018Implicit3O,Tremblay2018TrainingDN,Tobin2017DomainRF}, our method learns a neural network to perform random style transformation based on the style information with the extra random noise. 

Recently, some works in domain randomization tasks \cite{Yue2019DomainRA, Huang2021FSDRFS} aim to improve the generalization ability of the model based on the learnable network, which are relevant to our work. Specifically, \cite{Yue2019DomainRA} proposes to randomize the synthetic images with the styles of real images using auxiliary datasets based on the trained CycleGAN. And \cite{Huang2021FSDRFS} achieves domain randomization through converting spatial images into multiple frequency components and randomizing images in frequency space. Both of the two methods perform randomization and augmentation on \textit{data-level}, while the goal of our method it to learn a simple yet effective domain random network on \textit{feature-level}, which is a more goal-oriented and sample-diverse way.
	
\subsection{Style Transfer}
Style transfer is to translate the style of an image to the style of another image while preserving its content at the same time. Gatys \textit{et al.}~\cite{Gatys2016ImageST} demonstrate impressive style transfer results by matching Gram matrices of the neural activations from different convolutional layers. Li
\textit{et al.}~\cite{Li2017DemystifyingNS} propose a novel style loss
which aligns the convolutional feature statistics (\ie, mean and standard deviation) of the feature maps between stylized images and generated images. AdaIN~\cite{Huang2017ArbitraryST} 
achieves arbitrary style transfer in real-time by replacing the convolutional feature statistics of the content image with
those of the style image. Different from the image style transfer, our task is to learn domain-invariant feature representation via feature-based style randomization for seen source domains and unseen target domains. 

Similar methods leveraging AdaIN to augment style statistics have already been explored in prior domain adaptation and generalization works \cite{luo2020adversarial,wang2021learning}. Specifically, \cite{luo2020adversarial} generates the stylized images through a pre-trained Random Adaptive Instance Normalization (RAIN) module and produces diverse styles iteratively with adversarial style mining. \cite{wang2021learning} produces novel style images by mixing several transformations, which are produced by replacing the scaling and shifting statistics of AdaIN with the learnable parameters. However, the generated
styles of these methods are limited. For example, in \cite{luo2020adversarial}, since the target domain is known, the generated
domain distribution is guided based on the anchor style. And for \cite{wang2021learning}, the style statistics are generated by mixing numerous learnable parameters, which will overfit the
source domain because of lacking additional information. Different from these works, our FSR can perform random style transformation based on the style information with the extra random noise, ensuring that each perturbation can produce different and meaningful styles, which can generate the features with diverse styles
to enhance the generalization ability of the network.

\subsection{Adversarial Attacking}

Adversarial attacking aims to increase the robustness of the model by generating adversarial samples based on feature-level \cite{volpi2018adversarial, yang2020adversarial} or image-level \cite{huang2021rda}. These perturbed examples are designed for fooling machine learning models in various ways, \ie, fast gradient signs \cite{goodfellow2014explaining}, minimal adversarial perturbation \cite{moosavi2016deepfool}, universal adversarial perturbations \cite{moosavi2017universal}, transferable adversarial sample generation \cite{liu2019transferable}, \etc. For example, \cite{volpi2018adversarial} proposes a Conditional GAN to perform data augmentation on feature-level and learn domain-invariant features for unsupervised domain adaptation issue. \cite{yang2020adversarial} shares similar spirit with \cite{volpi2018adversarial}, while they investigate adversarial training in the field of semantic segmentation to generate point-wise perturbations. Besides, \cite{huang2021rda} designs an innovative Fourier adversarial attacking technique to generate adversarial samples.

Departing from the imperceptible attacks considered in adversarial training, we aim to learn models that are resistant to larger perturbations, namely out-of-distribution samples. The goal of the augmented samples in most existing adversarial attacking methods is commonly to let the discriminator be unable to distinguish the perturbed data from the original data. But they lack explicit constraints on image style information. Such the generated adversarial samples are not fully applicable to the issue of domain generalization as diverse domain variations are not captured. Instead, our method generates new features with random styles by the proposed FSR and employs a domain classifier to enhance the diversity of generated style information.

\begin{figure*}[h]
\centering
\includegraphics[width=.95\linewidth]{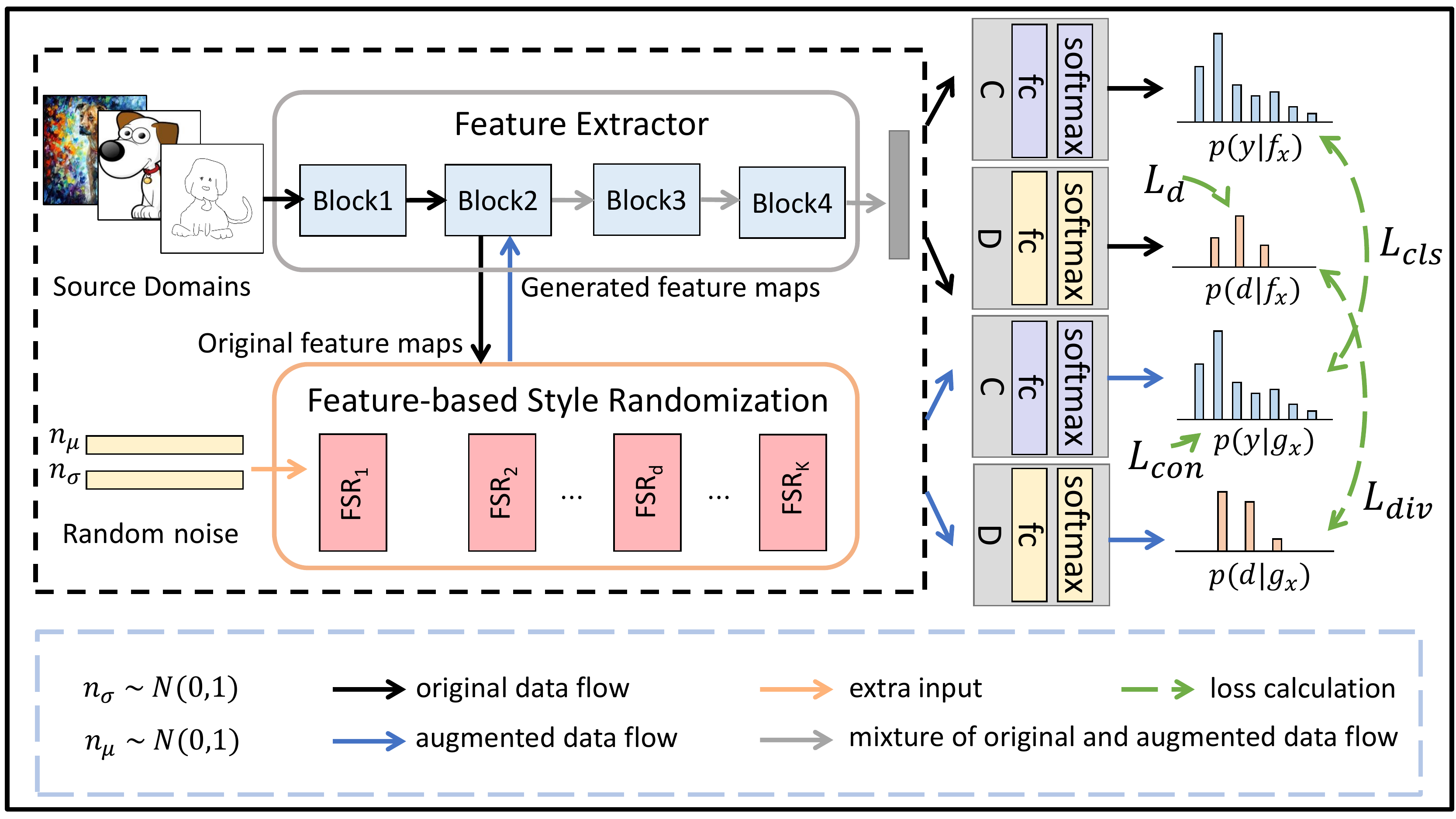}
\caption{An overview of our network.We take ResNet-18 as an example, which consists of 4 blocks. We apply the proposed FSR module over each block and show the process of inserting behind the second block here. The number of source domains is K, and each source domain is assigned an independent FSR. The outputs of the FSR are sent back to the current position of the feature extractor, to obtain the augmented stylized feature representations. Besides,
we introduce a domain discriminator to give an additional regularization to FSR, which aims to distinguish the augmented styles and original styles. Note that the parameters of two domain discriminators denoted by ``D'' and two classifiers denoted by ``C'' are shared.}
\label{fig:short}
\end{figure*}

\section{The proposed method}
	
	In this section, we first analyze our problem and discuss the preliminaries about our work. Then, we provide the detail of our proposed feature-based style randomization. Moreover, the architecture of our model is presented. Finally, we discuss our proposed progressive training scheme.
	
	Particularly, DG resorts to train the model on multiple source domains $\mathbf{D}_{S}=\lbrace D_{1},D_{2},\ldots,D_{K} \rbrace$ and generalizes to an arbitrary unseen target domain $D_{T}$ without any fine-tuning or re-training steps. Note that all of them share the same label space but have different data distributions.
	To attack the key issue, we propose the Feature-based Style Randomization (FSR) module to generate novel styles for each source domain. Taking advantage of AdaIN \cite{Huang2017ArbitraryST} developed to handle the style transfer task, we propose to generate new styles through randomly perturbing the style statistics (\ie, channel-wise mean and variance) of the input features. 
	Moreover, we put forward a novel progressive training strategy to ensure that all parameters of the model are fully trained. 
	An overview of our network is illustrated in Fig.~\ref{fig:short}. We present the proposed method in detail in the following parts.
	
	\subsection{Preliminaries}
	It has been known that the convolutional feature statistics of CNN can represent the style information of an image, such as the second-order statistics or channel-wise mean and variance \cite{Gatys2016ImageST,Li2016CombiningMR,Li2017DemystifyingNS}. Based on the style statistics, Ulyanov \textit{et al.} \cite{ulyanov2017improved} propose instance normalization (IN) to remove image style in the style transfer model. Given an input training image $x$, the feature maps of $x$ can be defined as $f_{x} \in \mathbb{R}^{C\times H\times W}$, where $H$ and $W$ indicate spatial dimensions, and $C$ is the number of channels. Thus, IN can be formulated as:
	\begin{equation}
	{\rm IN}(f_x)= \gamma \frac{f_x-f_\mu}{f_\sigma}+\beta,
	\end{equation}
	where $\gamma,\beta \in \mathbb{R}^{C}$ are learnable affine transformation parameters, and $f_{\mu},f_{\sigma} \in \mathbb{R}^{C}$  represent the channel-wise mean and standard deviation of each feature map:
	\begin{equation}
	f_{\mu}=\frac{1}{HW}\sum_{h=1}^{H}\sum_{w=1}^{W}f_{chw},
	\label{eq:7}
	\end{equation}
	\begin{equation}
	f_{\sigma}=\sqrt{\frac{1}{HW}\sum_{h=1}^{H}\sum_{w=1}^{W}(f_{chw}-f_{\mu})^2 + \epsilon},
	\label{eq:8}
	\end{equation}
	where $\epsilon$ is a constant for numerical stability. Furthermore, based on the style statistics, Huang \textit{et al.} \cite{Huang2017ArbitraryST} develop AdaIN to convert the style of images into a specific style, which replaces the affine parameters by the specific style statistics (\ie, $(s_\mu, s_\sigma)$). AdaIN can be formulated as follows:
    \begin{equation}
        {\rm AdaIN}(f_x,s)=s_\sigma \frac{f_x-f_\mu}{f_\sigma}+s_\mu.
    \end{equation}

    In this paper, we will perturb the statistics (\ie, the aforementioned channel-wise mean and standard deviation of each feature map) to achieve the style randomization. Also, we leverage AdaIN to replace the original style information by generated random style statistics.

	\subsection{Feature-based Style Randomization}
\begin{figure}[t]
		\begin{center}
			\includegraphics[width=1\linewidth]{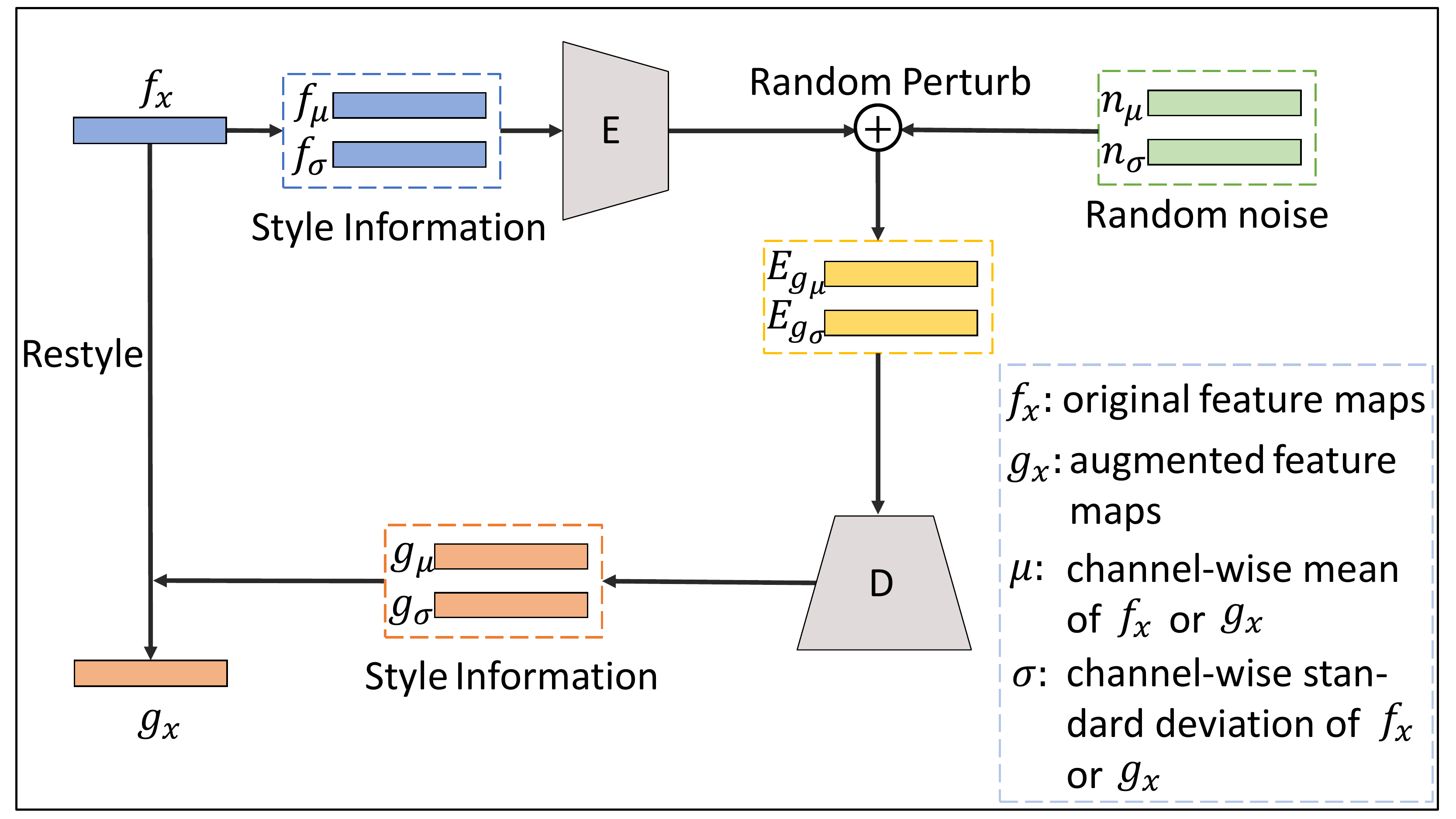}
		\end{center}
		\caption{An overview of our Feature-based Style Randomization (FSR) network. ``E'' and ``D'' denote encoder and decoder, respectively.}
		\label{fig:2}
	\end{figure}

	In this paragraph, we describe the architecture of FSR, as illustrated in Fig.~\ref{fig:2}. The proposed module can generate diverse styles to achieve the feature-based augmentation by the learned network. As indicated in \cite{Huang2017ArbitraryST}, we adopt the channel-wise mean and standard deviation $f_{\mu},f_{\sigma}$ in Eq. (\ref{eq:7}) and Eq. (\ref{eq:8}) as the style information. Particularly, we employ the encoder-decoder network to integrate the noise information into the original style in a hidden space. 

	Concretely, since the style information of an image are extracted from different channels, we introduce an encoder, consisting of a FC layer $\Theta_a$ and a ReLU layer, to mine the meaningful style information and the correlation among different channels. 
	We sample two noise variables $n_{\mu}$ and $n_{\sigma}\sim N(0, 1)$ with the same dimension as the original feature style embeddings to generate random style embeddings. Afterwards, we can perturb the original style statistics in the embedding space by a non-linear way. Thus, disturbed feature style embeddings $E_{g_\mu},E_{g_\sigma} \in \mathbb{R}^D$ can be written as:
	\begin{equation}
	E_{g_{\mu}}=\alpha\cdot {\rm ReLU}\big(\Theta_a(f_{\mu})\big)+(1-\alpha)\cdot n_{\mu},
	\label{eq:9}
	\end{equation}
	\begin{equation}
	E_{g_{\sigma}}=\alpha\cdot {\rm ReLU}\big(\Theta_a(f_{\sigma})\big)+(1-\alpha)\cdot n_{\sigma},
	\label{eq:10}
	\end{equation}
	where $\alpha$ is a random number sampled from Uniform(0, 1). Then we decode the perturbed embeddings into the original space by a decoder, including a FC layer $\Theta_b$ and a ReLU layer, to obtain the statistics of style information. Finally, we re-style the original feature maps $f_{x}$ by the generated novel style statistics to get the new augmented feature maps $g_{x}$ as:
	\begin{equation}
	g_{x} = {\rm ReLU}\big(\Theta_b(E_{g_{\sigma}})\big)\cdot(\frac{f_{x}-f_{\mu}}{f_{\sigma}})+{\rm ReLU}\big(\Theta_b(E_{g_{\mu}})\big).
	\label{eq:11}
	\end{equation}

	\subsection{The Architecture of Our Network}
	We now introduce the architecture and loss function of our network.
	In order to learn a promising augmentation scheme, we propose Feature-based Style Randomization (FSR), which can perform random style conversion with random noise as shown in Fig.~\ref{fig:2}. Since the proposed module is learned and executed in the feature space, diverse and abstract transformations of input images can be achieved, thus the augmented features are expected to cover more possible styles or distributions compared with image based augmentation. It is worth noting that the style statistics in different domains have different distributions, which leads the independent network for each domain being necessary, \ie, each domain has its private FSR. In this meaning, each domain can learn the individual augmentation strategy that suits itself. The architecture of our network is illustrated in Fig.~\ref{fig:short}.
	
	The general DG framework consists of a feature extractor $F_{f}$ and a classifier $F_{c}$. Our goal is to train the feature extractor $F_{f}$ that focuses on domain-independent semantic information by feature-based style randomization. Our proposed FSR can take place in different positions of the feature extractor, which could produce different influences for the final learned model. We will discuss the position selection in Section~\ref{sec3.4}. For each input of the network $(x,y,d)$, where $d$ and $y$ are the domain label and class label of the sample $x$, respectively, the feature maps from the $i$-th layer of $F_{f}$ is denoted as ${f_{x}^i}$, 
	and we denote the rest layers after the $i$-th layer of the feature extractor as ${F^r_{f}}$. We sample two noise variables $n_{\mu}$ and $n_{\sigma}\sim N(0, 1)$. The augmented stylized feature maps of the $i$-th layer of $F_{f}$ through domain-specific FSR can be thus written as:
	\begin{equation}
	{g_{x}^i}= {\rm FSR}_d({f_{x}^i},n_{\mu},n_{\sigma}).
	\label{eq:1}
	\end{equation}
	
	
	Our network, as shown in Fig.~\ref{fig:short}, is composed by three different modules (\ie, \underline{domain discriminator}, \underline{FSR}, and \underline{feature extractor and classifier}). We optimize the three components utilizing three different losses (\ie, $L_{d}$, $L_{\rm{fsr}}$ and $L_{\rm{cls}}$), respectively, which are described in the following part.
	
	We hope that the augmented features could have different distributions or style information from existing source domains. Thus we introduce a domain discriminator $F_{d}$ to address this problem as in~\cite{Zhou2020DeepDI}, which could effectively distinguish features of different domains. To be specific, $F_{d}$ is trained on source domains, which can be defined as:
	\begin{equation}
	L_{d}=L_{\rm{CE}}\big(F_{d}(F^r_{f}(f_{x}^i)),d\big),
	\label{eq:3}
	\end{equation}
	where $L_{\rm{CE}}$ is the cross-entropy loss.

	Then, we can train FSR by maximizing the loss of domain discriminator, which could further encourage FSR to explore unseen style information so as to improve the diversity of augmented features. The diversity loss of FSR is defined as:
	\begin{equation}
	L_{\rm{div}}=-L_{\rm{CE}}\big(F_{d}(F^r_{f}(g_{x}^i)),d\big).
	\label{eq:4}
	\end{equation}
	
	Besides, in order to ensure that the augmented features maintain the original semantic information, we introduce semantic consistency constraint $L_{\rm{con}}$ to assign the transformed features to original classes, which can be written as:
	\begin{equation}
	L_{\rm{con}}=L_{\rm{CE}}\big(F_{c}(F^r_{f}(g_{x}^i)),y\big).
	\label{eq:2}
	\end{equation}
	Therefore, the total loss of FSR can be written as:
	\begin{equation}
	L_{\rm{fsr}}=L_{\rm{con}}+L_{\rm{div}}.
	\label{eq:5}
	\end{equation}
	
	At last, we use both the augmented features and the original features to train the feature extractor and the classifier by the cross-entropy loss:
	\begin{equation}
	L_{\rm{cls}}=L_{\rm{CE}}\big(F_{c}(F^r_{f}(f_{x}^i)),y\big)+\lambda \cdot L_{\rm{CE}}\big(F_{c}(F^r_{f}(g_{x}^i)),y\big),
	\label{eq:6}
	\end{equation}
	where $\lambda$ is the hyper-parameter to trade off the two classification losses of original features and augmented features.

    \begin{figure}
    \centering
    \subfigure[PACS]{
    \includegraphics[width=4.1cm]{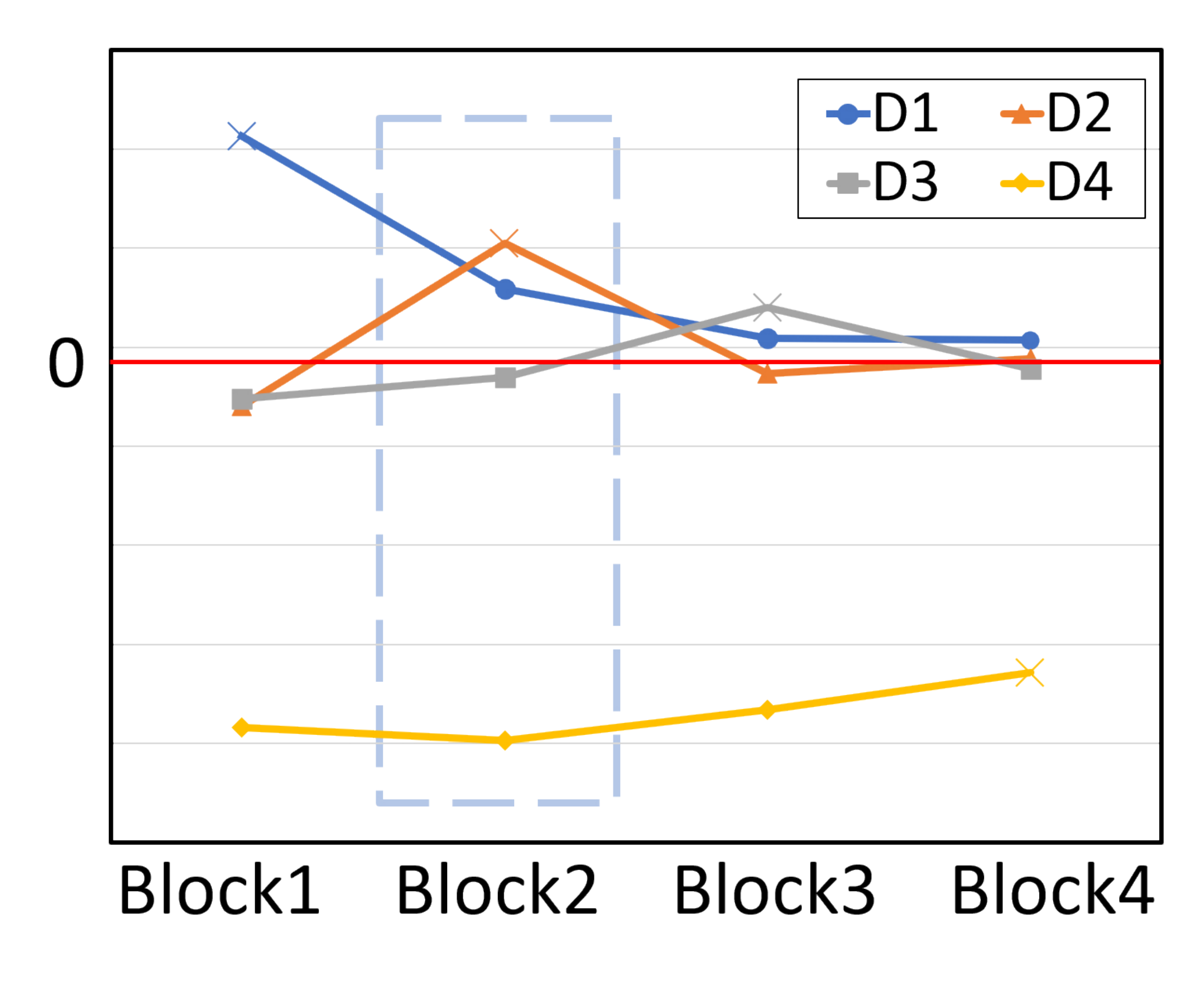}
    }
    \subfigure[OfficeHome]{
    \includegraphics[width=4.1cm]{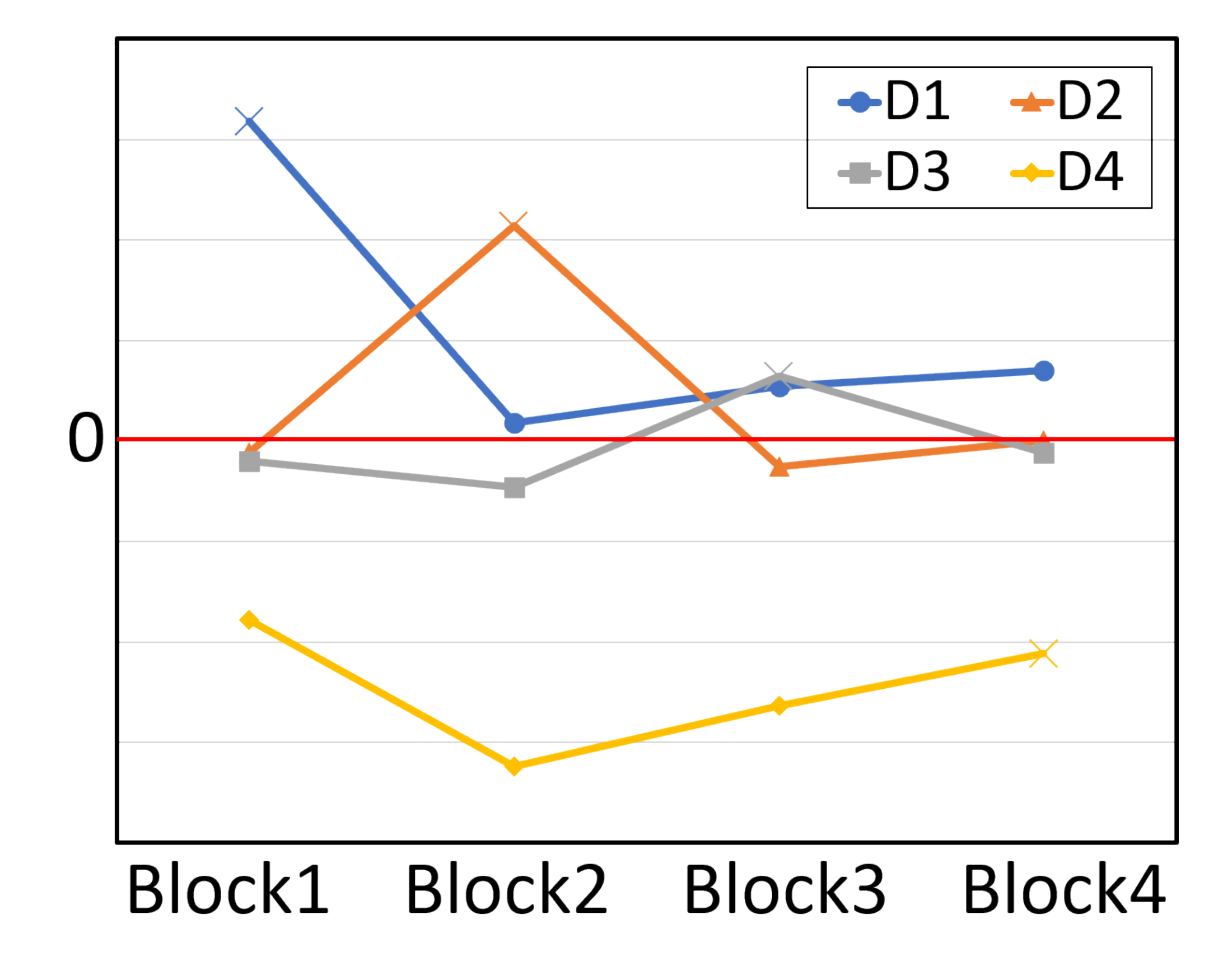}
    }
    \caption{The effectiveness of FSR when plugging into different positions of ResNet-18. The horizontal axis represents the position of plugging FSR, and the vertical axis is $d_{fsr}^i-d_{bal}^i$ from the $i$-th block of ResNet-18 (\ie, D$i$).}
    \label{fig:4}
    \end{figure}

	\subsection{Progressive Training Scheme} ~\label{sec3.4}
	In order to investigate the influence of inserting the proposed FSR module into different positions of the feature extractor, 
	we calculate the domain discrepancy via Eq.  (\ref{eq:12}). Specifically, taking ResNet-18 as an example, which contains four residual blocks denoted by \texttt{Block1-4}, we obtain the domain discrepancy according to the features from the $i$-th block  in ResNet-18 as follows:
	\begin{equation}
	d^i = \frac{2}{K(K-1)}\sum^{K}_{m=1}\sum^{K}_{n=m+1}|{\rm GAP}(\overline{f}^{i}_{m}) - {\rm GAP}(\overline{f}^{i}_{n})|,
	\label{eq:12}
	\end{equation}
	where $\overline{f}^{i}_{m}$ represents the averaged feature maps of all samples from the $i$-th block of the feature extractor in the $m$-th domain, $K$ is the number of source domains and $\rm{GAP}(.)$ is the global average pooling operation. Then, we calculate the domain difference $d_{\rm{fsr}}^i$ according to features from the $i$-th block during using our FSR module. 
	Also, we obtain the domain difference $d_{\rm{bal}}^i$ for the baseline model.
	
	It is worth noting that, the ``negative'' value $d_{\rm{fsr}}^i-d_{\rm{bal}}^i$ shows the effectiveness of our module, \ie, if the result is negative, it indicates that using the FSR can reduce the domain discrepancy, as shown in Fig.~\ref{fig:4}. In this figure, (a) shows the results on PACS when considering Art Painting as the target domain and use the other three domains to train the model, and (b) is the results on OfficeHome when considering RealWorld as the target domain.
	We can conclude the following observations:
	
	\begin{enumerate}
	 \item The domain discrepancy based on the block with FSR will be large when compared to the baseline. For example, ``D2'' is greater than 0 in the dotted box of Fig.~\ref{fig:4} (a). This confirms the proposed FSR can indeed enhance the style diversity. 
	 \item FSR has an obvious positive effect on the subsequent network architecture, which encourage them to focus on domain-agnostic information. For example, ``D3'' and ``D4'' is smaller than 0 in the dotted box of Fig.~\ref{fig:4} (a). Therefore, the low-level noise can help to learn the high-level domain-invariant features.
	\end{enumerate}
	
	Based on these observations, we propose to progressively add the proposed FSR into each block in a front-to-back order. 
	
	\setlength{\textfloatsep}{0.1cm}
	\begin{algorithm}[t]
		\caption{Training procedure of our network.}
		\label{alg:A}
		\begin{algorithmic}[1]
			\REQUIRE Source domains $D_s$, feature extractor $F_f$, label classifier $F_c$, domain classifier $F_d$, domain-specific FSR $F_{fsr}$, the number of training stages $N$, max iteration of each training stage $T$.\\
			\ENSURE Learned feature extractor $F_f$ and classifier $F_c$.\\
			\STATE Insert $F_{fsr}$ behind the first block of $F_f$.\\
			\FOR {$n=1$ to $N$} 
			\FOR {$t=1$ to $T$}
			\STATE $(x,y,d) \sim D_s$.\\
			\STATE Obtain original features $f_x$ and stylized features $g_x$.\\
			\STATE Update $F_d$ using Eq. (\ref{eq:3}) .\\
			\STATE Update $F_{fsr}$ using Eq. (\ref{eq:5}).\\
			\STATE Update $F_c$, $F_f$ using Eq. (\ref{eq:6}).\\
			\STATE Obtain trained $F^1_f$, $F^1_c$, $F^1_{fsr}$, $F^1_d$. \\
			\ENDFOR
			\STATE Move $F_{fsr}$ to the back of the next block of $F_f$. \\
			\ENDFOR
		\end{algorithmic}
	\end{algorithm}
	\setlength{\textfloatsep}{0.5cm}
	
	To be specific, firstly, we plug the FSR into the back of the first residual block \texttt{Block1} to train the model via $T$ iterations. Then we move FSR to the back of \texttt{Block2}, and utilize the model parameters trained in the previous stage as the initialization parameters to train the model via $T$ iterations. Similarly, the module is continuously moved until the last residual block \texttt{Block4}, thus we get the final trained model which consists of a feature extractor and a label classifier. 
	The overall training procedure is presented in Algorithm \ref{alg:A}. 
	

\section{Experiments}
In this section, we firstly introduce the experimental datasets and settings in Section~\ref{es1}. Then, we compare the proposed method with the state-of-the-art domain generalization methods on three standard benchmark datasets in Sections ~\ref{es2}. Furthermore, to validate the effectiveness of various components and the training scheme in the proposed framework, we conduct ablation studies in Section~\ref{es3}. Lastly, we further analyze the property of the proposed network and give the visualization results in Section~\ref{es4}.

	\subsection{Experimental Settings}\label{es1}
	\textbf{Datasets.} We evaluate our approach on three public DG benchmark datasets. 
	
	\begin{itemize}
	    \item \textbf{PACS} \cite{Li2017DeeperBA} consists of four different domains: Photo, Art, Cartoon and Sketch. It contains 9,991 images with 7 object categories in total.
	    \item \textbf{Office-Home} \cite{venkateswara2017deep} contains around 15,500 images of 65 categories of office and home objects. It has four different domains namely Art, Clipart, Product and Real, which is originally introduced for UDA but is also applicable in the DG setting.
	    \item \textbf{VLCS} \cite{torralba2011unbiased} contains images with 5 categories collected from four different datasets: PASCAL VOC 2007 \cite{everingham2007pascal}, LabelMe \cite{Russell2007LabelMeAD}, Caltech-101 \cite{fei2004learning} and Sun09 \cite{choi2010exploiting}, in which all images are photos captured with different cameras or compositions. 
	\end{itemize}
	
	\textbf{Implementation details.} Following \cite{carlucci2019domain}, we divide the dataset into training set and validation set according to 9:1 in the case of PACS and OfficeHome, and 7:3 in the case of VLCS. We use any three of them as source domains to train the model and test on the left one. We select the model with the best result on the validation set for test, and the final reported result is the average of three runs. We choose ResNet-18 pretrained on ImageNet as our backbone following~\cite{li2019episodic,carlucci2019domain}, and adopt the hyperparameters proposed in~\cite{carlucci2019domain}. Specifically, we set the batch size to 128, and train the model using SGD with a momentum of 0.9, a weight decay of $5e-4$. The initial learning rate of the convolutional and BN layer of the backbone is $1e-3$, and the rest of the network is $1e-2$, respectively. The network is trained for a total of 60 epochs, where every 15 epochs is considered as a training stage with the same learning rate. We simply aggregate all source domains to train the model via 60 epochs as baseline model. The value of hyperparameter $\lambda$ is 1 for PACS and VLCS, and 0.8 for OfficeHome, respectively.

	\subsection{Comparison with Other Methods}\label{es2}
	\renewcommand{\cmidrulesep}{0mm} 
    \setlength{\aboverulesep}{0mm} 
    \setlength{\belowrulesep}{0mm} 
    \setlength{\abovetopsep}{0cm}  
    \setlength{\belowbottomsep}{0cm}
    
		\begin{table}
		\renewcommand\arraystretch{1.3}
		\begin{center}
		\caption{Comparison with the state-of-the-art methods (\%) on PACS using ResNet-18 as backbone. The title of each column indicates the name of unseen target domain, and the best result of each target domain is marked in bold. 
		} 
		\label{tab:1}
			\begin{tabular}{l|ccccc}
			    \specialrule{0.8pt}{2pt}{2pt}
				Method & Art & Cartoon & Photo & Sketch & Avg\\
				
				\specialrule{0pt}{1pt}{1pt}
				\hline
				\specialrule{0pt}{1pt}{1pt}
			
				JiGen~\cite{carlucci2019domain} & 79.42 & 75.25 & 96.03 & 71.35 & 80.51 \\	CrossGrad~\cite{Shankar2018GeneralizingAD} & 79.80 & 76.80 & 96.03 & 71.35 & 80.70\\
				SFA~\cite{li2021simple}& 81.20 & 77.80 & 93.90 & 73.70 & 81.70\\
				Epi-FCR~\cite{li2019episodic} & 82.10 & 77.00 & 93.90 & 73.00 & 81.50\\
				L2A-OT~\cite{Zhou2020LearningTG} & 83.30 & 78.20 & 96.20 & 73.60 & 82.80\\
				DDAIG~\cite{Zhou2020DeepDI} & 84.20 & 78.10 & 95.30 & 74.70 & 83.10\\
				SagNet~\cite{nam2021reducing} & 83.58 & 77.66 & 76.30 & 95.47 & 83.25\\
				MixStyle~\cite{zhou2021domain} & 84.10 & 78.80 & 96.10 & 75.90 & 83.70\\
				NAS-OoD~\cite{bai2021ood}& 83.74 &  79.69 & 77.27 & \textbf{96.23} & 84.23\\

				FACT~\cite{xu2021fourier} & \textbf{85.37} & 78.38 & 95.15 & 79.15 & 84.51\\
				DSON~\cite{Seo2019LearningTO} & 84.57 & 76.82 & 95.51 & 81.83 & 84.68\\
				
				\specialrule{0pt}{1pt}{1pt}
				\hline
				\specialrule{0pt}{1pt}{1pt}
				
				Our Method & 84.49 & \textbf{81.15} & 96.13 & \textbf{82.01} & \textbf{85.95}\\
				\specialrule{0.8pt}{2pt}{2pt}
			\end{tabular}
		\end{center}
	
	\end{table}
	
	\begin{table}
	\renewcommand\arraystretch{1.3}
		\begin{center}
		\caption{Comparison with the state-of-the-art domain generalization methods (\%) on OfficeHome using a ResNet-18 as CNN backbone. For details about the meaning of columns and marks, refer to Table \ref{tab:1}.}
		\label{tab:2}
			
				\begin{tabular}{l|ccccc}
				\specialrule{0.8pt}{2pt}{2pt}
					Method & Art & Clipart & Product & Real & Avg\\
    				\specialrule{0pt}{1pt}{1pt}
    				\hline
    				\specialrule{0pt}{1pt}{1pt}
					JiGen~\cite{carlucci2019domain} & 53.04 & 47.51 & 71.47 & 72.79 & 61.20\\
					SagNet~\cite{nam2021reducing} & 60.20 & 45.38 & 70.42& 73.38 & 62.34\\
					DSON~\cite{Seo2019LearningTO} & 59.29 & 45.74 & 71.60 & 73.51 & 62.53\\	CrossGrad~\cite{Shankar2018GeneralizingAD} & 58.40 & 49.40 & 73.90 & 75.80 & 64.40\\
					DDAIG~\cite{Zhou2020DeepDI} & 59.20 & 52.30 & 74.60 & 76.00 & 65.50\\
					MixStyle~\cite{zhou2021domain} & 58.70 & 53.40 & 74.20 & 75.90 & 65.50\\
					L2A-OT~\cite{Zhou2020LearningTG} & \textbf{60.60} & 50.10 & 74.80 & \textbf{77.00} & 65.60 \\
					\specialrule{0pt}{1pt}{1pt}
    				\hline
    				\specialrule{0pt}{1pt}{1pt}
					Our Method & 59.95 & \textbf{55.07} & \textbf{74.82} & 76.34 & \textbf{66.55}\\
					\specialrule{0.8pt}{2pt}{2pt}
			\end{tabular}
		\end{center}
		
	\end{table}
	
We compare our method with the recent state-of-the-art domain generalization methods. We choose seven methods based on data augmentation which are the most relevant to our method, including DVEN \cite{Liu2021DomainGV}, FACT \cite{xu2021fourier}, SFA \cite{li2021simple}, DDAIG \cite{Zhou2020DeepDI}, L2A-OT \cite{Zhou2020LearningTG}, CrossGrad \cite{Shankar2018GeneralizingAD} and MixStyle \cite{zhou2021domain}. And there are five methods based on learning domain-independent feature representations, including SagNet~\cite{nam2021reducing}, DSON \cite{Seo2019LearningTO}, MMD \cite{Li2018DomainGW}, D-SAM \cite{DInnocente2018DomainGW} and TF \cite{Li2017DeeperBA}. Besides, the meta-learning-based methods are also compared with our method, including NAS-OoD \cite{bai2021ood}, Epi-FCR \cite{li2019episodic} and JiGen \cite{carlucci2019domain}. We conduct the comparative experiments on all three datasets.

The results on PACS are reported in Table~\ref{tab:1}. As observed, our method largely outperforms the state-of-the-art method (\ie, DSON) which develops domain-specific normalization based on batch and instance normalizations to capture domain-independent representations. Note that DDAIG \cite{Zhou2020DeepDI}, L2A-OT \cite{Zhou2020LearningTG}, CrossGrad \cite{Shankar2018GeneralizingAD} are all image-based-augmentation methods, and among them, our method shows the best performance when evaluated on the unseen domains. In particular, compared with L2A-OT \cite{Zhou2020LearningTG}, the state-of-the-art image-based-augmentation method by synthesizing diverse images with a conditional generative network, our method obtains a gain of 2.85\% (85.95\% vs. 83.10\%).  And for FACT~\cite{xu2021fourier} which augmented images based on fourier methods, our method is 1.44\% (85.95\% vs. 84.51\%) better, which owes to our feature-based augmentation is more goal-oriented than image-based augmentation. Compared to MixStyle \cite{zhou2021domain}, which simply mixes the statistics on feature-level, there is an improvement of 2.25\% (85.95\% vs. 83.70\%). Besides, for SFA~\cite{li2021simple} which also induced random noise to perturb original features, our method gains 4.25\% (85.95\% vs. 81.70\%). It illustrates that our proposed encoder-decoder-based learnable perturbing network and progressive training scheme could produce more diverse styles. Furthermore, our method has a great improvement on the two difficult domains, \ie, Cartoon and Sketch. 

We also conduct experiments on OfficeHome. It is worth noting that OfficeHome has a relatively smaller domain shift and larger data amount than PACS. As shown in Table~\ref{tab:2}, our feature-based-augmentation method can achieve the best performance compared to other methods. Similarly, our method improves significantly on Clipart with a large domain gap by 1.67\% (55.07\% vs. 53.40\%) when compared with MixStyle \cite{zhou2021domain}, the state-of-the-art method on Clipart. 

Since most of existing methods tested on VLCS use AlexNet as the backbone, we also choose AlexNet for a fair comparison. However, due to the few number of convolutional layers in AlexNet, we only conduct a simple experiment. Specifically, we regard AlextNet as a block, and insert FSR after the last convolutional layer for training. Table~\ref{tab:3} reports the experimental results. Although we can observe that VLCS has a much smaller domain shift than PACS and OfficeHome, our method still achieves considerable improvement compared with the state-of-the-art method JiGen~\cite{carlucci2019domain}. Note that although DVEN~\cite{Liu2021DomainGV} also adopts the Gaussian noise base perturbation for feature generation, our method advances it via directly and domain-specific layer-wise style perturbation. As observed in Table~\ref{tab:3}, our method has an obvious improvement of 3.77\% (75.95\% vs. 72.18\%) on average accuracy. Furthermore, we will take ResNet-18 as the backbone to perform ablation study on VLCS to further illustrate the effectiveness of the proposed method in the next part.

	\begin{table}
	\renewcommand\arraystretch{1.3}
		\begin{center}
		\caption{Comparison with the state-of-the-art methods (\%) on VLCS using AlexNet as backbone. For details about the meaning of columns and marks, refer to Table \ref{tab:1}.}
		\label{tab:3}
			\footnotesize{
				\begin{tabular}{l|ccccc}
					\specialrule{0.8pt}{2pt}{2pt}

					Method & Caltech & Labelme & Pascal & Sun & Avg\\
					
					\specialrule{0pt}{1pt}{1pt}
    				\hline
    				\specialrule{0pt}{1pt}{1pt}
					
					D-SAM~\cite{DInnocente2018DomainGW} & 91.75 & 56.95 & 58.59 & 60.84 & 67.03 \\
					MMD~\cite{Li2018DomainGW} & 94.40 & 62.60 & 67.70 & 64.40 & 72.28 \\
					TF~\cite{Li2017DeeperBA} & 93.63 & 63.49 & 69.99 & 61.32 & 72.11\\
					DVEN~\cite{Liu2021DomainGV} & 91.52 & \textbf{64.68} & 65.58 & 66.94 & 72.18\\
					JiGen~\cite{carlucci2019domain} & 96.93 & 60.90 & 70.62 & 64.30 & 73.19\\
					SFA~\cite{li2021simple}& 97.20 & 62.00 & 70.40 & 66.20 & 74.00\\
					\specialrule{0pt}{1pt}{1pt}
    				\hline
    				\specialrule{0pt}{1pt}{1pt}
					Our Method & \textbf{97.95} & 61.03 &\textbf{71.94} & \textbf{71.42} & \textbf{75.59}\\
					\specialrule{0.8pt}{2pt}{2pt}
			\end{tabular}}
		\end{center}
	\end{table}
	
	\subsection{Ablation Study}\label{es3}
	
	\begin{table}
	\renewcommand\arraystretch{1.3}
		\begin{center}
		\caption{Ablation study (\%) on PACS based on ResNet-18 to verify the effectiveness of each component of the proposed method. }
		\label{tab:5}
			\footnotesize{
				\begin{tabular}{l|ccccc}
					\specialrule{0.8pt}{2pt}{2pt}
    	
					Method & Art & Cartoon & Photo & Sketch & Avg\\
					
					\specialrule{0pt}{1pt}{1pt}
    				\hline
    				\specialrule{0pt}{1pt}{1pt}
					
					Baseline & 77.23 & 75.16 & 95.51 & 69.67 & 79.39\\
					w/o $L_{\rm{div}}$ & 81.70 & 79.31 & 95.41 & 77.51 & 83.49 \\
					w/o $L_{\rm{con}}$ & 79.49 & 78.67 & 94.93 & 77.57 & 82.67\\
					w/o spec. & 80.78 & 78.46 & 94.89 & 78.47 & 83.15\\
					w/o noise & 80.42 & 78.55 & 95.75 & 76.22 & 82.74\\
					w/o E\&D & 18.57 & 16.60 & 11.32 & 19.65 & 16.54 \\
					\specialrule{0pt}{1pt}{1pt}
    				\hline
    				\specialrule{0pt}{1pt}{1pt}
					
					Our Method & \textbf{84.49} & \textbf{81.15} & \textbf{96.13} & \textbf{82.01} & \textbf{85.95}\\
					\specialrule{0.8pt}{2pt}{2pt}
    				
			\end{tabular}}
		\end{center}
		
	\end{table}

\subsubsection{Evaluation of each component of FSR}
In this part, we conduct the experiment on the PACS dataset to validate the efficacy of each component in our network, as shown in Table~\ref{tab:5}, where ``w/o $L_{\rm{div}}$'' is the removal of diversity loss in Eq. (\ref{eq:3}), ``w/o $L_{\rm{con}}$'' is the removal of consistency loss in Eq. (\ref{eq:4}), ``w/o spec.'' means that we share the feature-based style randomization network for all source domains, and ``w/o noise'' denotes we only feed the original feature maps into the random style transform network without introducing additional noise. Moreover, ``w/o E\&D" denotes removing the learnable parameters of our FSR network namely the encoder and decoder networks. Baseline is simply using all source domains to train the baseline model (\ie, ``Deepall'').
	
As indicated in Table~\ref{tab:5}, our method consistently outperforms all variants. First, our method improves the accuracy by 2.46\% (85.95\% vs. 83.49\%) over ``w/o $L_{\rm{div}}$'', which shows that using domain-guidance optimization for style augmentation can ensure the diversity of the augmented distributions. Second, the result of ``w/o $L_{\rm{con}}$'' will drop by 3.28\% (82.67\% vs. 85.95\%), which validates that the semantic consistency loss could preserve the semantic information of original images. Besides, the result without learning an independent FSR network for each source domain (\ie, ``w/o spec.'') decreases by 2.80\% (83.15\% vs. 85.95\%). It could indicate that domain-specific FSR could help for the sample-diversity, which is more suitable for DG. Furthermore, we validate the effectiveness of introducing extra noise. As observed in Table~\ref{tab:5}, our method has a 3.21\% (85.95\% vs. 82.74\%) improvement compared with ``w/o noise'', which shows that although directly converting the style information of the original feature maps through the learnable network can augment the style, the extra random noise will introduce more diverse styles or distributions into the model at the same time, thus it is also critical to enhance the generalization ability of the model. Also, we conduct the experiment by adding the random noise to the style information directly without the encoder and decoder networks as indicated by ``w/o E\&D". The performance is inferior, and what's worse, the training fails to converge, leading the results in each target domain almost being equal to random prediction. The reason is easy to interpret---directly adding noise without any constraint will destroy the semantic information and style information of the original features, which could easily cause the network to become unstable and collapse. Therefore, this results further validate the effectiveness of our FSR network.

\subsubsection{Evaluation of training scheme}
In this section, we evaluate the effectiveness of the proposed training scheme on PACS, VLCS and OfficeHome dataset, respectively. 
	
\textbf{Results on PACS.} The experimental results are reported in Table~\ref{tab:ab1}. The first line represents the results of the baseline model, which is trained using all source domains based on the basic backbone. The multiple checkmarks (\ie, ``$\checkmark$'') in a row represent training the network by gradually changing the position of FSR according to our training strategy. In addition, we also make an experiment in which FSR is applied over all blocks of the feature extractor at the same time (\ie, four ``$\star$'' in a row). As shown in Table~\ref{tab:ab1}, compared with the baseline model, applying FSR over each block of the feature extractor can improve the accuracy, especially inserting it after the first block can improve the accuracy by 3.58\% (82.97\% vs. 79.39\%). Besides, adding our FSR module over each block of the feature extractor at the same time reduces the result by 4.34\% (75.05\% vs. 79.39\%) when compared to baseline. Specifically, the performance can be further improved using our training scheme. The second stage is improved by 0.69\% (83.66\% vs. 82.97\%) compared with the first stage, the third stage is further improved by 0.41\% (84.07\% vs. 83.66\%) , and the last stage further increases the result by 1.88\% (85.95\% vs. 84.07\%).
	
	\begin{table}[t]
		\renewcommand\arraystretch{1.3}
		\begin{center}
			\caption{Ablation study (\%) on PACS based on ResNet-18 to verify the effectiveness of our progressive training scheme. The top reports the results of applying FSR over each individual block. The bottom shows the results of each training stage when inserting FSR into the back of each block progressively. Blo.1-4 represent four residual blocks of the ResNet-18. The asterisk ($\star$) indicates simultaneously applying FSR over all blocks from the beginning of the training stage.}
			\label{tab:ab1}
			\footnotesize{
				\begin{tabular}{p{0.45cm}<{\centering}p{0.45cm}<{\centering}p{0.45cm}<{\centering}p{0.45cm}<{\centering}|p{0.50cm}<{\centering}p{0.70cm}<{\centering}p{0.58cm}<{\centering}p{0.58cm}<{\centering}c}
				\specialrule{0.8pt}{2pt}{2pt}
    				\multicolumn{4}{c|}{Position} & \multicolumn{5}{c}{PACS} \\
    				\cmidrule(lr){1-4} \cmidrule(lr){5-9}
					Blo.1 & Blo.2 & Blo.3 & Blo.4 & Art & Cartoon & Photo & Sketch & Avg\\
					\specialrule{0pt}{1pt}{1pt}
    				\hline
    				\specialrule{0pt}{1pt}{1pt}
					$-$ & $-$ & $-$ & $-$ & 77.23 & 75.16 & 95.51 & 69.67 & 79.39 \\
					$\star$ & $\star$ & $\star$ & $\star$ & 72.31 & 72.95 & 91.32 & 63.60 & 75.05\\
					$\checkmark$ & $-$ & $-$ & $-$ & 80.99 & 76.42 & 94.73 & 79.72 & 82.97\\
					$-$ &$\checkmark$ &$ -$ & $-$ & 80.43 & 75.80 & 94.05 & 81.49 & 82.95\\
					$-$ & $-$ &$\checkmark$ & $-$ & 79.45 & 76.47 & 95.23 & 71.25 & 80.59\\
					$-$ & $-$ & $-$ &$\checkmark$ & 80.26 & 77.22 & 96.09 & 71.47 & 81.26\\
					\specialrule{0pt}{1pt}{1pt}
					\hline
					\specialrule{0pt}{1pt}{1pt}
					$\checkmark$ & $\checkmark$ & $-$ & $-$ & 80.06 & 77.60 & 94.21 & 82.75 & 83.66 \\
					$\checkmark$ & $\checkmark$ & $\checkmark$ & $-$ & 81.61 & 78.83 & 94.61 & 81.23 & 84.07 \\
					$\checkmark$ & $\checkmark$ & $\checkmark$ & $\checkmark$ & \textbf{84.49} & \textbf{81.15} & \textbf{96.13} & \textbf{82.01} & \textbf{85.95} \\
					\specialrule{0.8pt}{2pt}{2pt}
    				
			\end{tabular}}
		\end{center}
		
	\end{table}
	
\textbf{Results on VLCS.} Similar to the results on PACS, the results on VLCS have the same rules, as seen in Table \ref{tab:ab2}. FSR also has a positive effect in each position on the VLCS dataset, especially inserting it after the first and the last block can improve the accuracy by 0.99\% (75.35\% vs. 74.36\%). Adding FSR module into each block of the feature extractor at the same time will break the performance by 4.52\% (69.84\% vs. 74.36\%). The accuracy can be gradually improved according to our proposed training scheme, and finally reached a 2.26\% (76.62\% vs. 74.36\%) improvement based on the baseline model.

	\begin{table}[t]
		\renewcommand\arraystretch{1.3}
		\begin{center}
			\caption{Ablation study (\%) on VLCS based on ResNet-18 to verify the effectiveness of our progressive training scheme. For details about the meaning of columns and marks, refer to Table~\ref{tab:ab1}. }
			\label{tab:ab2}
			\footnotesize{
			\begin{tabular}{p{0.45cm}<{\centering}p{0.45cm}<{\centering}p{0.45cm}<{\centering}p{0.45cm}<{\centering}|p{0.58cm}<{\centering}p{0.76cm}<{\centering}p{0.58cm}<{\centering}p{0.58cm}<{\centering}c}
				\specialrule{0.8pt}{2pt}{2pt}	
				\multicolumn{4}{c|}{Position} & \multicolumn{5}{c}{VLCS} \\
					\cmidrule(lr){1-4} \cmidrule(lr){5-9}
					Blo.1 & Blo.2 & Blo.3 & Blo.4 & Caltech & Labelme & Pascal & Sun & Avg\\
					\specialrule{0pt}{1pt}{1pt}
					\hline
					\specialrule{0pt}{1pt}{1pt}
					$-$ & $-$ & $-$ & $-$ & 95.85 & 58.01 & 71.96 & 71.63 & 74.36 \\
					$\star$ & $\star$ & $\star$ & $\star$ & 86.20 & 56.97 & 66.36 & 69.81 & 69.84\\
					$\checkmark$ & $-$ & $-$ & $-$ & 96.18 & 58.65 & 73.75 & 72.83 & 75.35\\
					$-$ &$\checkmark$ &$ -$ & $-$ & 96.52 & 60.00 & 73.98 & 70.44 & 75.23\\
					$-$ & $-$ &$\checkmark$ & $-$ & 94.05 & 59.68 & 73.59 & 72.72 & 75.01\\
					$-$ & $-$ & $-$ &$\checkmark$ & 96.52 & 59.89 & \textbf{74.51} & \textbf{72.83} & 75.35\\
					\specialrule{0pt}{1pt}{1pt}
					\hline
					\specialrule{0pt}{1pt}{1pt}
					$\checkmark$ & $\checkmark$ & $-$ & $-$ & 96.18 & 60.16 & 73.56 & 71.61 & 75.38 \\
					$\checkmark$ & $\checkmark$ & $\checkmark$ & $-$ & 96.30 & 61.88 & 73.72 & 71.70 & 75.90 \\
					$\checkmark$ & $\checkmark$ & $\checkmark$ & $\checkmark$ & \textbf{96.95} & \textbf{63.99} & 73.65 & 71.90 & \textbf{76.62} \\
					\specialrule{0.8pt}{2pt}{2pt}
			\end{tabular}}
		\end{center}
	\end{table}
	
\textbf{Results on OfficeHome.} As the results on PACS and VLCS, we can notice the results on OfficeHome obey the same law as shown in Table~\ref{tab:ab3}. Applying the FSR module over each position at the same time will damage the accuracy, but applying FSR over each block separately will improve the generalization performance of the model, especially adding it over the last block will increase the accuracy by 2.39\% (65.19\% vs. 62.80\%). Our proposed training scheme is still effective on OfficeHome. Specifically, applying FSR over each position gradually according to our training scheme will make the result have a progressive improvement, and finally reach a result that is 3.70\% (66.50\% vs. 62.80\%) better than the baseline model.
	
	\begin{table}[t]
		\renewcommand\arraystretch{1.3}
		\begin{center}
			\caption{Ablation study (\%) on OfficeHome based on ResNet-18 to verify the effectiveness of our progressive training scheme. For details about the meaning of columns and marks, refer to Table~\ref{tab:ab1}.}
			\label{tab:ab3}
			\footnotesize{
			\begin{tabular}{p{0.45cm}<{\centering}p{0.45cm}<{\centering}p{0.45cm}<{\centering}p{0.45cm}<{\centering}|p{0.58cm}<{\centering}p{0.68cm}<{\centering}p{0.68cm}<{\centering}p{0.58cm}<{\centering}c}
				\specialrule{0.8pt}{2pt}{2pt}	\multicolumn{4}{c|}{Position} & \multicolumn{5}{c}{OfficeHome} \\
					\cmidrule(lr){1-4} \cmidrule(lr){5-9}
					Blo.1 & Blo.2 & Blo.3 & Blo.4 & Art & Clipart & Product & Real & Avg\\
					\specialrule{0pt}{1pt}{1pt}
					\hline
					\specialrule{0pt}{1pt}{1pt}
					$-$ & $-$ & $-$ & $-$ & 57.17 & 49.03 & 71.91 & 73.07 & 62.80 \\
					$\star$ & $\star$ & $\star$ & $\star$ & 54.93 & 46.32 & 68.19 & 69.09 & 59.63\\
					$\checkmark$ & $-$ & $-$ & $-$ & 57.16 & 50.33  & 72.27 & 73.81 & 63.39\\
					$-$ &$\checkmark$ &$ -$ & $-$ & 57.76 & 51.38 & 72.19 & 73.21 & 63.64\\
					$-$ & $-$ &$\checkmark$ & $-$ & 58.37 & 52.87 & 72.51 & 73.29 & 64.26\\
					$-$ & $-$ & $-$ &$\checkmark$ & 58.23 & 54.32 & 73.17 & 75.05 & 65.19\\
					\specialrule{0pt}{1pt}{1pt}
					\hline
					\specialrule{0pt}{1pt}{1pt}
					$\checkmark$ & $\checkmark$ & $-$ & $-$ & 57.80 & 51.89 & 72.67 & 73.98 & 64.09 \\
					$\checkmark$ & $\checkmark$ & $\checkmark$ & $-$ & 58.44 & 54.05 & 73.78 & 74.12 & 65.10 \\
					$\checkmark$ & $\checkmark$ & $\checkmark$ & $\checkmark$ & \textbf{59.95} & \textbf{55.07} & \textbf{74.59} & \textbf{76.34} & \textbf{66.50} \\
					\specialrule{0.8pt}{2pt}{2pt}
			\end{tabular}}
		\end{center}
	\end{table}
	
Based on the above experimental results, we can conclude that 1) using FSR in each block can obtain better performance when compared to the baseline, which confirms the effectiveness of FSR. 2) Adding our FSR module over each block of the feature extractor at the same time cannot enhance the generalization ability and will deteriorate the performance to some extent. This is because our method executes on the feature level and introduces noise for randomization, thus it will bring too much randomness to the network. Moreover, the noise destroys model stability, which results in performance degradation. 3) Our proposed training strategy can gradually improve performance with the progressive introduction of FSR, which means that the proposed training scheme indeed makes the network be available to more possible distributions, so that the model could have good generalization ability in unseen domains.

\subsection{Further Analysis}\label{es4}
	
	\begin{table}[t]
		\renewcommand\arraystretch{1.3}
		\begin{center}
			\caption{Experimental results (\%) on PACS based on ResNet-34. For details about the meaning of columns and marks, refer to Table~\ref{tab:ab1}.}
			\label{tab:s1}
			\footnotesize{
				\begin{tabular}{p{0.45cm}<{\centering}p{0.45cm}<{\centering}p{0.45cm}<{\centering}p{0.45cm}<{\centering}|p{0.58cm}<{\centering}p{0.70cm}<{\centering}p{0.58cm}<{\centering}p{0.58cm}<{\centering}c}
				\specialrule{0.8pt}{2pt}{2pt}	
				\multicolumn{4}{c|}{Position} & \multicolumn{5}{c}{PACS} \\
					\cmidrule(lr){1-4} \cmidrule(lr){5-9}
					Blo.1 & Blo.2 & Blo.3 & Blo.4 & Art & Cartoon & Photo & Sketch & Avg\\
					\specialrule{0pt}{1pt}{1pt}
    				\hline
    				\specialrule{0pt}{1pt}{1pt}
					$-$ & $-$ & $-$ & $-$ & 81.98 & 77.28 & 96.57 & 70.92 & 81.69 \\
					$\star$ & $\star$ & $\star$ & $\star$ & 76.86 & 78.91 & 95.93 & 74.94 & 81.66\\
					$\checkmark$ & $-$ & $-$ & $-$ & 84.47 & 77.44 & 96.43 & 79.07 & 84.35\\
					$-$ &$\checkmark$ &$ -$ & $-$ & 82.99 & 80.36 & 96.01 & 84.65 & 86.00\\
					$-$ & $-$ &$\checkmark$ & $-$ & 84.20 & 82.57 & \textbf{96.95} & 78.41 & 85.53\\
					$-$ & $-$ & $-$ &$\checkmark$ & 83.87 & 79.94 & 96.47 & 75.87 & 84.04\\
					\specialrule{0pt}{1pt}{1pt}
    				\hline
    				\specialrule{0pt}{1pt}{1pt}
					$\checkmark$ & $\checkmark$ & $-$ & $-$ & 86.10 & 80.45 & 95.23 & \textbf{86.78} & 87.14 \\
					$\checkmark$ & $\checkmark$ & $\checkmark$ & $-$ & 86.48 & 82.81 & 95.87 & 85.04 & 87.55 \\
					$\checkmark$ & $\checkmark$ & $\checkmark$ & $\checkmark$ & \textbf{86.82} & \textbf{83.79} & 96.67 & 84.56 & \textbf{87.96} \\
					\specialrule{0.8pt}{2pt}{2pt}
			\end{tabular}}
		\end{center}
		
	\end{table}
	
	\begin{table}[t]
	\renewcommand\arraystretch{1.3}
		\begin{center}
		\caption{Experimental results (\%) on PACS based on VGG11. For details about the meaning of columns and marks, refer to Table \ref{tab:s1}.}
		\label{tab:s2}
			\footnotesize{
				\begin{tabular}{ccc|ccccc}
					\specialrule{0.8pt}{2pt}{2pt}
					\multicolumn{3}{c|}{Position} & \multicolumn{5}{c}{PACS} \\
					\cmidrule(lr){1-3} \cmidrule(lr){4-8}
					Blo.1 & Blo.2 & Blo.3 & Art & Cartoon & Photo & Sketch & Avg\\
					\specialrule{0pt}{1pt}{1pt}
    				\hline
    				\specialrule{0pt}{1pt}{1pt}
					$-$ & $-$ & $-$ & 75.55 & 71.88 & \textbf{96.51} & 61.31 & 76.31 \\
					$\star$ & $\star$ & $\star$ & 75.76 & 73.17 & 96.05 & 63.48 & 77.11\\
					$\checkmark$& $-$ & $-$ & \textbf{80.79} & 73.55 & 94.59 & 70.77 & 79.93\\
					$-$ &$\checkmark$ & $-$ & 75.88 & 73.95 & 93.71 & 74.20 & 79.43\\
					$-$ & $-$ &$\checkmark$ & 75.81 & 73.91 & 95.67 & 68.05 & 78.36\\
					\specialrule{0pt}{1pt}{1pt}
    				\hline
    				\specialrule{0pt}{1pt}{1pt}
					$\checkmark$ & $\checkmark$ & $-$ & \textbf{80.79} & 73.63 & 94.51 & 76.60 & 81.38 \\
					$\checkmark$ & $\checkmark$ & $\checkmark$ & 79.69 & \textbf{76.84} & 95.22 & \textbf{76.98} & \textbf{82.18} \\
					\specialrule{0.8pt}{2pt}{2pt}
			\end{tabular}}
		\end{center}
	\end{table}

	\begin{figure*}
		\begin{center}
			\subfigure[]{
				\label{Fighp.sub.1}	\includegraphics[width=5.7cm]{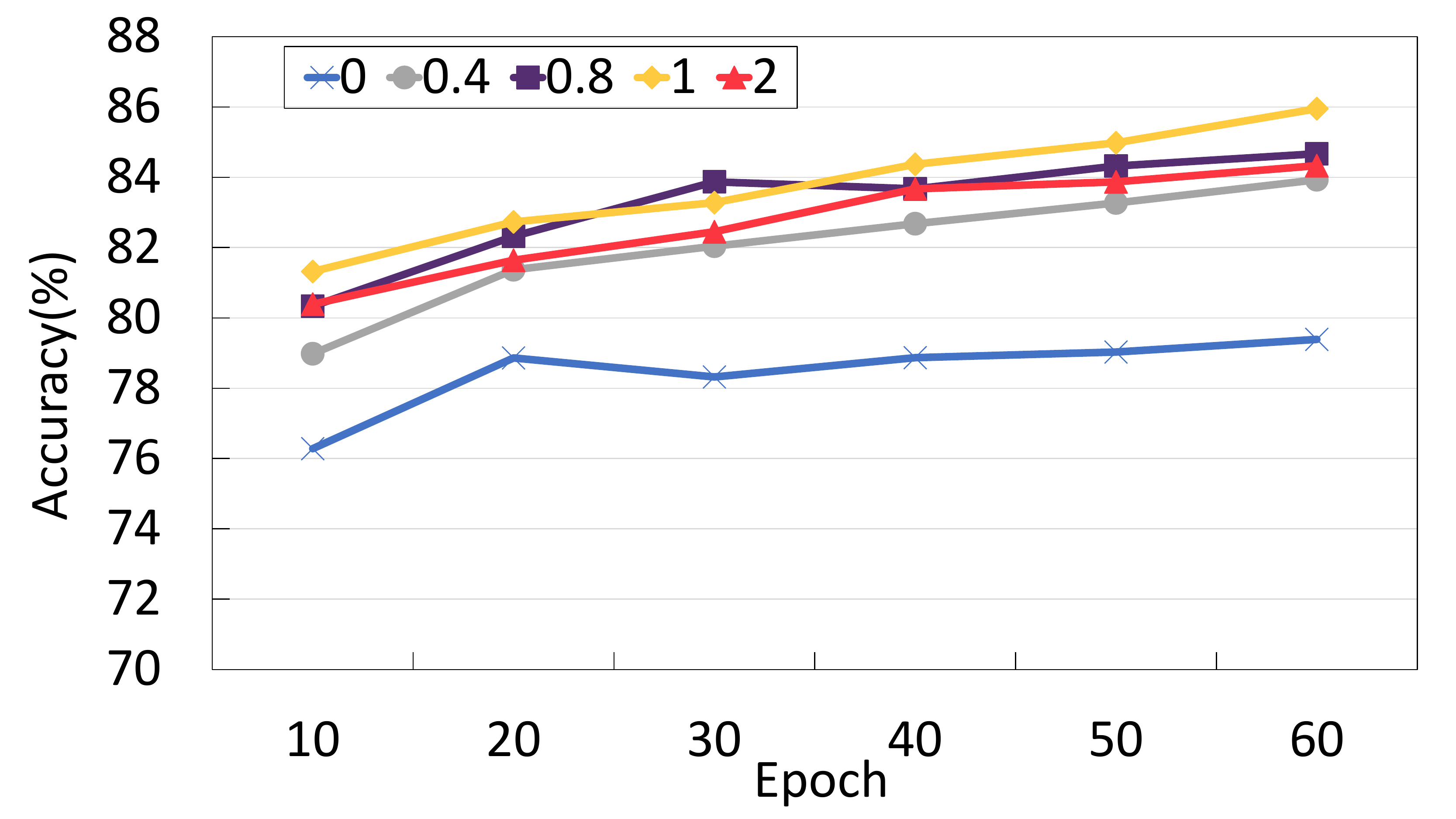}}
			\subfigure[]{
				\label{Fighp.sub.2}	\includegraphics[width=5.7cm]{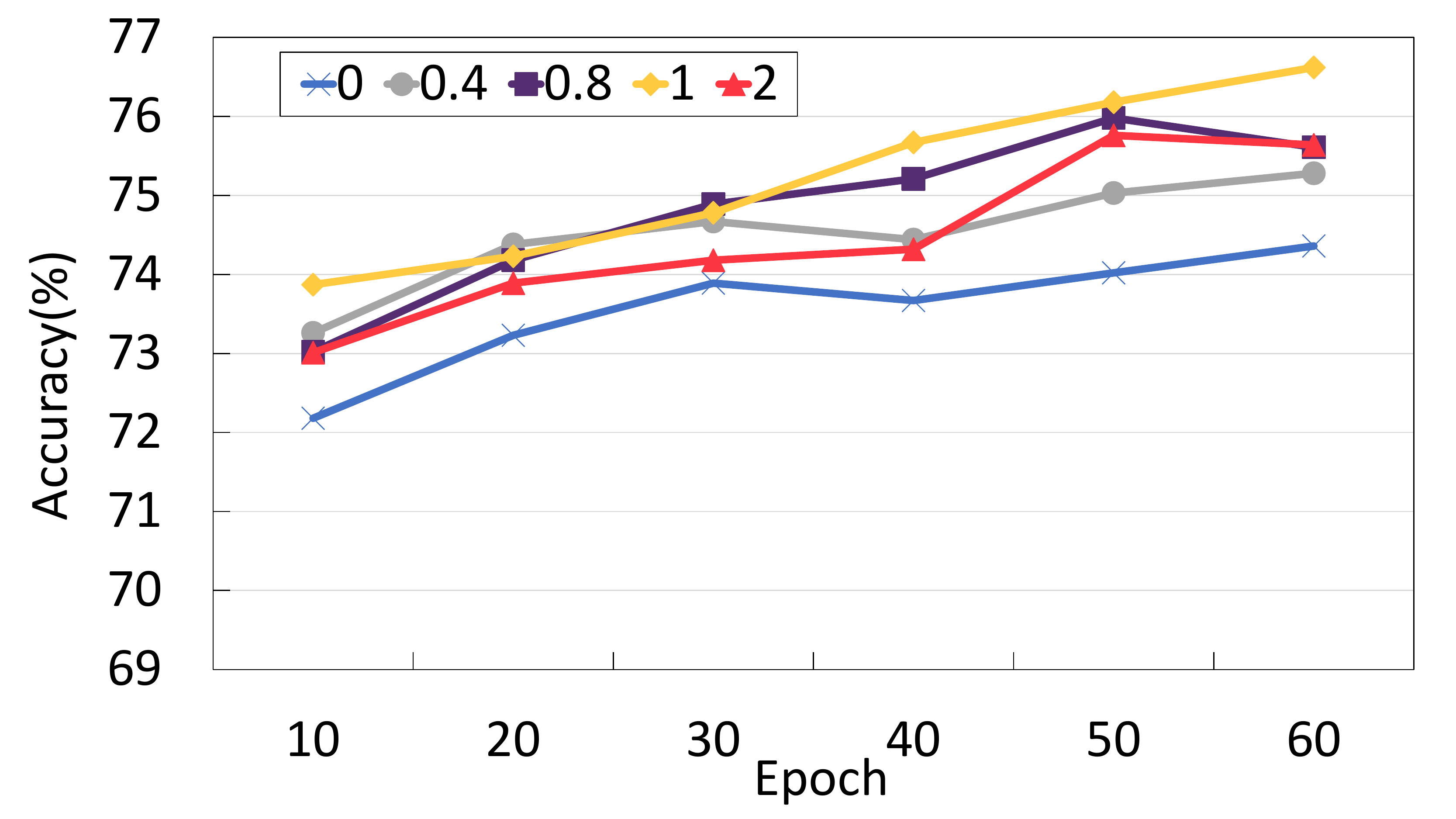}}
			\subfigure[]{
				\label{Fighp.sub.3}	\includegraphics[width=5.7cm]{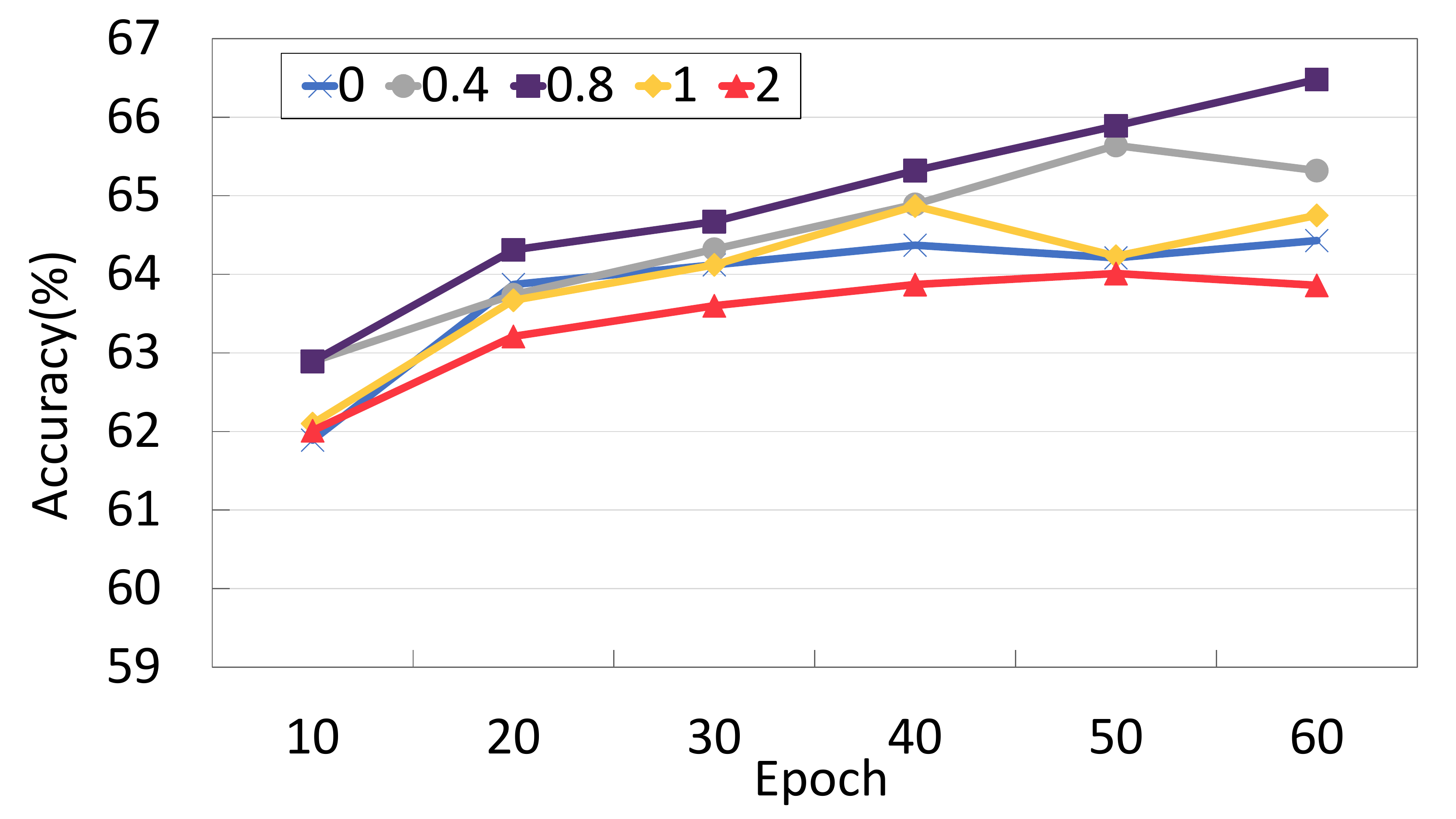}}
			\caption{Evaluation on the hyper-parameter $\lambda$ in Eq.~(\ref{eq:6}) on three standard benchmark datasets. (a), (b), and (c) represent the experimental results of PACS, VLCS and OfficeHome dataset, respectively.}
			\label{Fig:hp}
		\end{center}
	\end{figure*}

	\begin{table*}
	\renewcommand\arraystretch{1.3}
	\begin{center}
	\setlength{\tabcolsep}{4mm}
	\caption{Experimental results (\%) of different feature-based data manipulation methods with the progressive training scheme on PACS. The last column denotes the difference between the result of using our proposed progressive training scheme and the best result among the results of inserting the feature-based data manipulation into each position (\ie, the blue value in this table). ``w/o noise'' means the FSR module without introducing extra noise. For more details about the meaning of columns and marks, refer to Table \ref{tab:ab1}.}
	\label{tab:a1}
	\footnotesize{
		\begin{tabular}{c|cccc|ccccc|c}
				\specialrule{0.8pt}{2pt}{2pt}	
				\multirow{2}[1]{*}{ Method} & \multicolumn{4}{c|}{Position} &  \multicolumn{5}{c|}{PACS} & \multirow{2}[1]{*}{ Difference}\\
				\cmidrule(lr){2-5} \cmidrule(lr){6-10}
				& Blo.1 & Blo.2 & Blo.3 & Blo.4  & Art & Cartoon & Photo & Sketch & Avg & \\
				\specialrule{0pt}{1pt}{1pt}
				\cline{1-11}
				\specialrule{0pt}{1pt}{1pt}
				& $-$ & $-$ & $-$ & $-$ & 77.23 & 75.16 & 95.51 & 69.67 & 79.39 & $-$\\
				\specialrule{0pt}{1pt}{1pt}
				\hline
				\specialrule{0pt}{1pt}{1pt}
				\multirow{5}[1]{*}{\begin{sideways} MixStyle~\cite{zhou2021domain}\end{sideways}} & $\checkmark$ & $-$ & $-$ & $-$ & 79.82 & 76.41 & 96.31 & 68.70 & 80.31 &  \multirow{5}[1]{*}{$+0.68$}\\
				& $-$ &$\checkmark$ &$ -$ & $-$ & 81.66 & 77.32 & 96.29 & 70.82 & 81.52 &  \\
				& $-$ & $-$ &$\checkmark$ & $-$ & 80.18 & 77.77 & \textbf{97.01} & 73.36 & \textcolor{blue}{82.08} & \\
				&$-$ & $-$ & $-$ &$\checkmark$ & 79.91 & 75.57  & 95.79 & 68.70 & 79.99 & \\
				\specialrule{0pt}{1pt}{1pt}
				\cline{2-10}
				\specialrule{0pt}{1pt}{1pt}
				& \multicolumn{4}{c|}{train scheme} & \textbf{81.70} & \textbf{77.80} & 96.79 & \textbf{74.76} & \textbf{82.76} & \\
				\specialrule{0pt}{1pt}{1pt}
				\hline
				\specialrule{0pt}{1pt}{1pt}
				\multirow{5}[1]{*}{\begin{sideways} pAdaIN~\cite{Nuriel2020PermutedAR}\end{sideways}} & $\checkmark$ & $-$ & $-$ & $-$ & 80.53 & 76.51 & 96.03 & 69.63 & 80.67 & \multirow{5}[1]{*}{$+0.43$}\\
				& $-$ &$\checkmark$ &$ -$ & $-$ & \textbf{81.54} & 77.59 & 96.33 & 73.57 & 82.26 & \\
				& $-$ & $-$ &$\checkmark$ & $-$ & 80.55 & 77.02 & 96.63 & 76.00 & \textcolor{blue}{82.55} & \\
				&$-$ & $-$ & $-$ &$\checkmark$ & 77.16 & 72.23 & 92.93 & 64.75 & 76.78 & \\
				\specialrule{0pt}{1pt}{1pt}
				\cline{2-10}
				\specialrule{0pt}{1pt}{1pt}
				& \multicolumn{4}{c|}{train scheme} & 80.31 & \textbf{78.32} & \textbf{96.71} & \textbf{76.23} & \textbf{82.89} & \\
				\specialrule{0pt}{1pt}{1pt}
				\hline
				\specialrule{0pt}{1pt}{1pt}
				\multirow{5}[1]{*}{\begin{sideways} w/o noise\end{sideways}} & $\checkmark$ & $-$ & $-$ & $-$ & 77.71 & 74.65 & 95.15 & 70.57 & 79.52 & \multirow{5}[1]{*}{$+1.57$}\\
				& $-$ &$\checkmark$ &$ -$ & $-$ & 75.21 & 74.77 & 94.17 & 74.71 & 79.72 & \\
				& $-$ & $-$ &$\checkmark$ & $-$ & 78.06 & 71.13 & 95.23 & 65.23 & 77.41 & \\
				&$-$ & $-$ & $-$ &$\checkmark$ & 80.03 & 76.64 & \textbf{96.43} & 71.57 & \textcolor{blue}{81.17} & \\
				\specialrule{0pt}{1pt}{1pt}
				\cline{2-10}
				\specialrule{0pt}{1pt}{1pt}
				& \multicolumn{4}{c|}{train scheme} & \textbf{80.42} & \textbf{78.55} & 95.75 & \textbf{76.22} & \textbf{82.74} & \\
				\specialrule{0pt}{1pt}{1pt}
				\hline
				\specialrule{0pt}{1pt}{1pt}
				\multirow{5}[1]{*}{\begin{sideways} Our method\end{sideways}} & $\checkmark$ & $-$ & $-$ & $-$ & 80.99 & 76.42 & 94.73 & 79.72 & \textcolor{blue}{82.97} & \multirow{5}[1]{*}{\textbf{$+$2.98}}\\
				& $-$ &$\checkmark$ &$ -$ & $-$ & 80.43 & 75.80 & 94.05 & 81.49 & 82.95 & \\
				& $-$ & $-$ &$\checkmark$ & $-$ & 79.45 & 76.47 & 95.23 & 71.25 & 80.59 & \\
				&$-$ & $-$ & $-$ &$\checkmark$ & 80.26 & 77.22 & 96.09 & 71.47 & 81.26 & \\
				\specialrule{0pt}{1pt}{1pt}
				\cline{2-10}
				\specialrule{0pt}{1pt}{1pt}
				& \multicolumn{4}{c|}{train scheme} & \textbf{84.49} & \textbf{81.15} & \textbf{96.13} & \textbf{82.01} & \textbf{85.95} & \\
				\specialrule{0.8pt}{2pt}{2pt}
		\end{tabular}}
	\end{center}
\end{table*}
	
\textbf{Evaluation on different backbones.} We employ additional backbones to conduct the experiment on PACS dataset, which reveals that our proposed method is effective for different network architectures. Besides the ResNet-18 reported above, we use \textbf{ResNet-34} and \textbf{VGG11} as backbone. ResNet-34 includes four blocks, which is the same as ResNet-18. VGG11 consists of five stages, and we divide it into 3 blocks through 1-, 3-, and 5-th stages to ensure that the FSR module is not inserted too close. We apply FSR over each block to conduct the experiment on PACS. The experimental results are shown in Tables~\ref{tab:s1} and~\ref{tab:s2}. As observed, the similar conclusions can be obtained as using ResNet-18 as backbone (Table~\ref{tab:ab1} and Table~\ref{tab:ab2}). For different backbone networks, FSR also shows positive effect at each position. Specifically, inserting after the second block can increase 4.39\% (86.00\% vs. 81.69\%) for ResNet-34 and 3.62\% (79.93\% vs. 76.31\%) for VGG11. Besides, through our training strategy, the results could be further improved stage by stage. Compared with the best among the results applying FSR over each block, our training scheme increases from 86.00\% to 87.96\% for ResNet-34 and from 79.93\% to 82.18\% for VGG11, respectively. This sufficiently confirms the effectiveness of the proposed FSR and the progressive training scheme.

    \begin{table}[t]
	\renewcommand\arraystretch{1.3}
		\begin{center}
		\caption{Experimental results (\%) on the source domains of PACS using ResNet-18. Each column indicates the name of unseen target domain, and the result means the test accuracy on the held-out validation set.}
		\label{tab:source}
			\footnotesize{
				\begin{tabular}{l|ccccc}
					\specialrule{0.8pt}{2pt}{2pt}
					Method & Art & Cartoon & Photo & Sketch & Avg\\
					\specialrule{0pt}{1pt}{1pt}
					\hline
					\specialrule{0pt}{1pt}{1pt}
					Baseline & 96.90 & 96.12 & 95.44 & 96.21 & 96.17\\
					Our Method & \textbf{97.39} & \textbf{96.20} & \textbf{95.72} & \textbf{97.14} & \textbf{96.62}\\
					\specialrule{0.8pt}{2pt}{2pt}
			\end{tabular}}
		\end{center}
	\end{table}

	\begin{figure}
		\begin{center}
			\subfigure[Baseline]{
				\label{Fig5.sub.1}	\includegraphics[width=4.0cm]{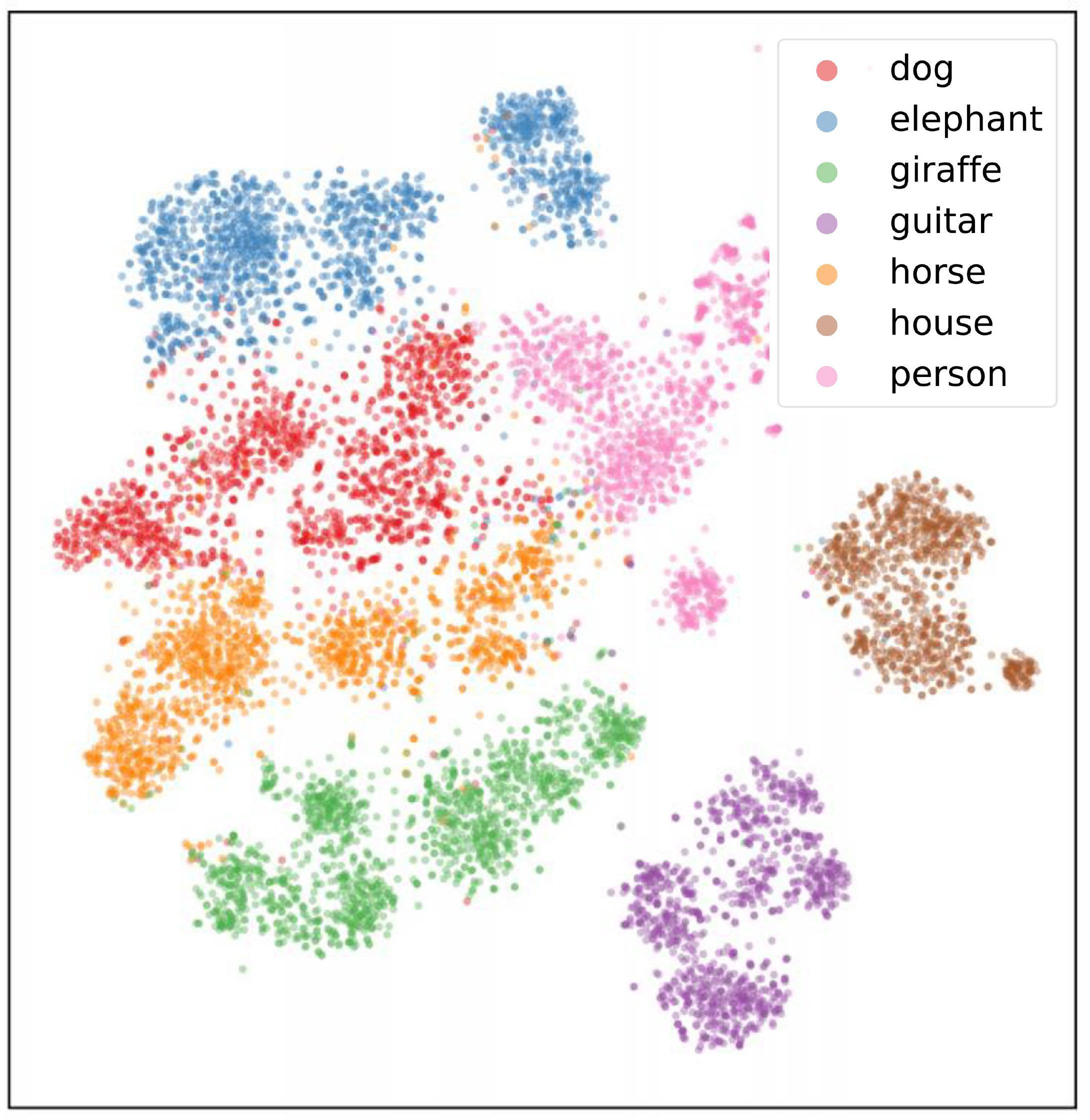}}
			\subfigure[Ours]{
				\label{Fig6.sub.1}	\includegraphics[width=4.0cm]{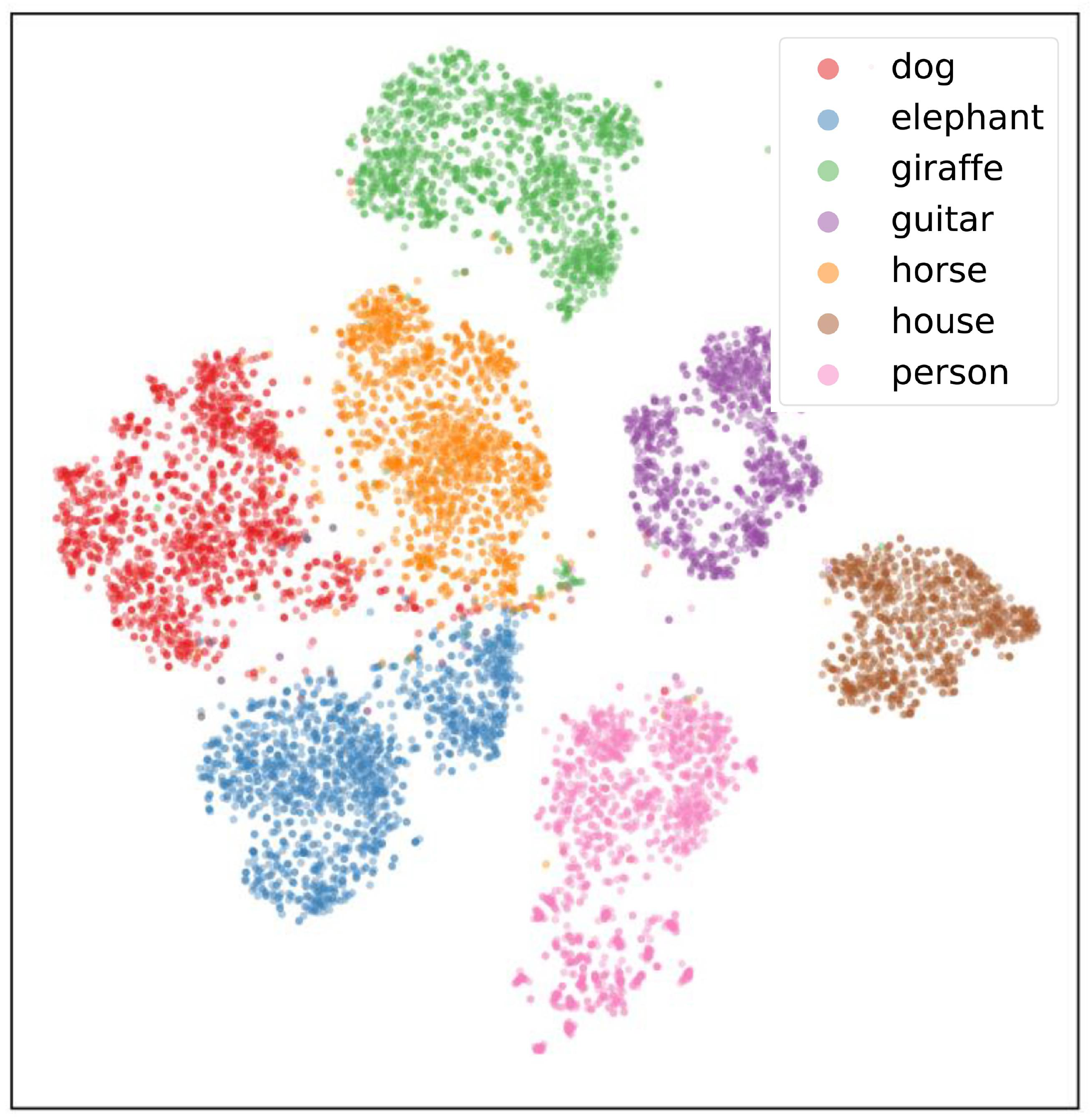}}
			\subfigure[Baseline]{
				\label{Fig5.sub.2}	\includegraphics[width=4.0cm]{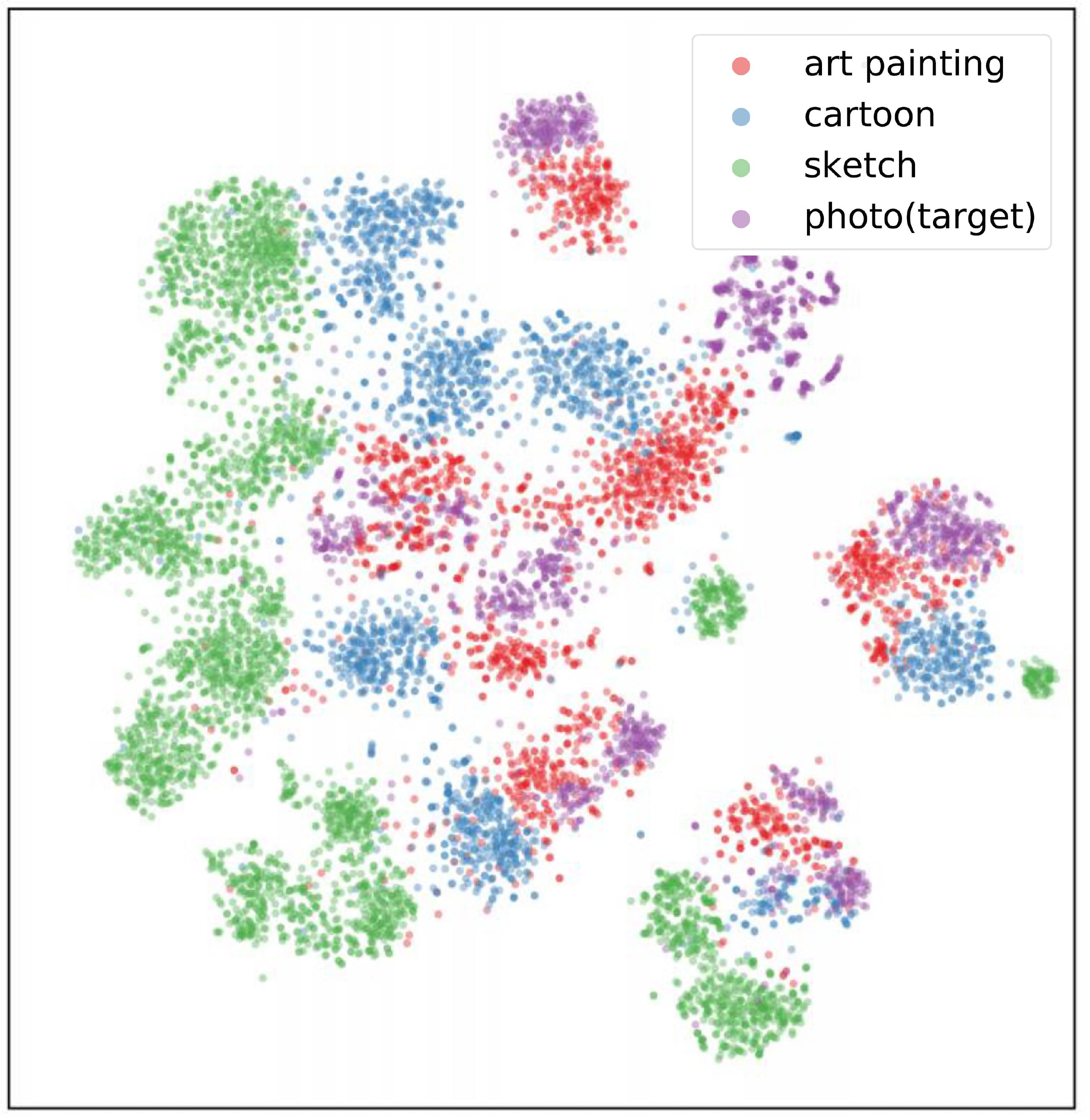}}
			\subfigure[Ours]{
				\label{Fig6.sub.2}	\includegraphics[width=4.0cm]{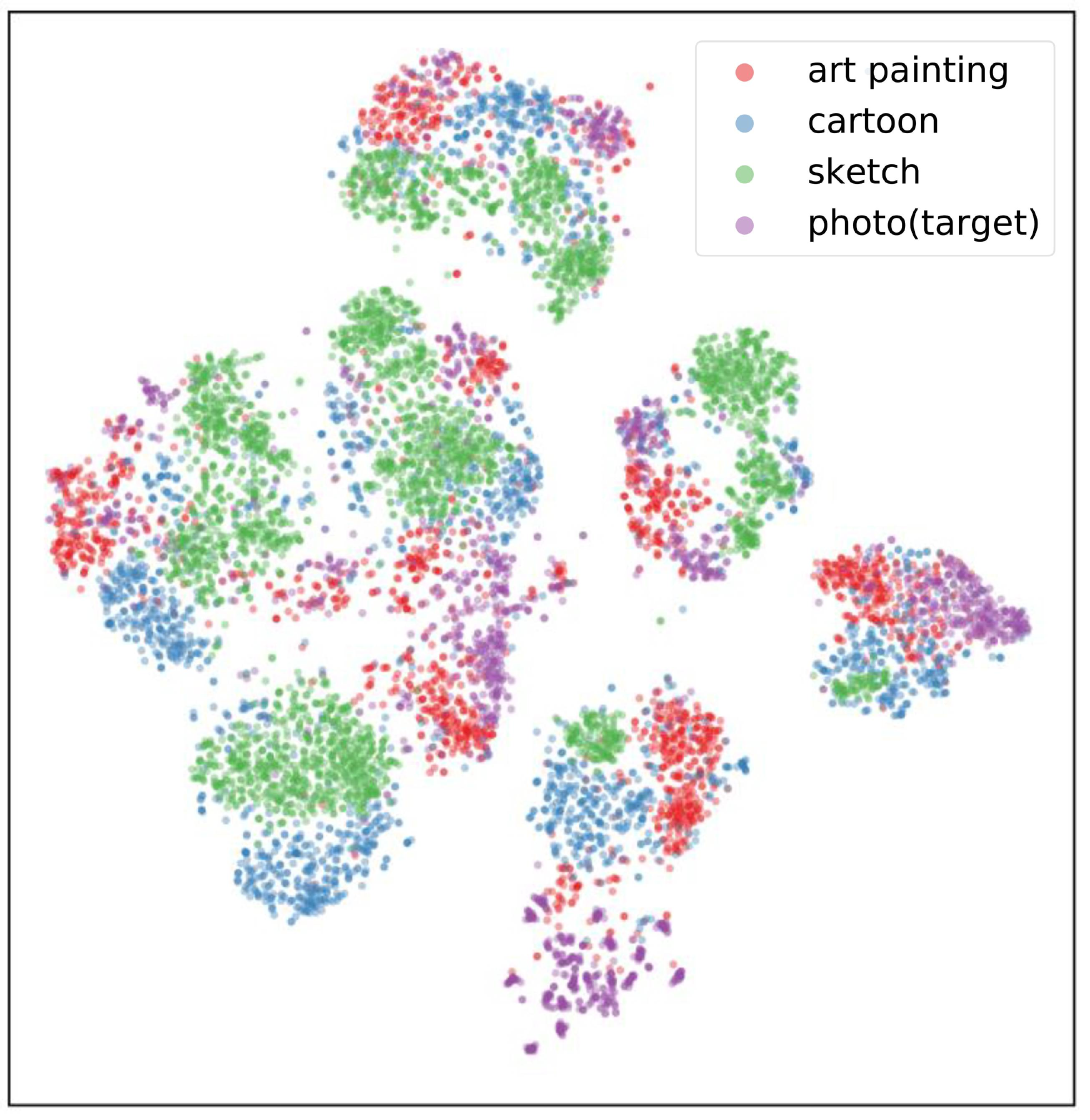}}
			\caption{The t-SNE \cite{Maaten2008VisualizingDU} visualization of feature representations extracted by the feature extractor of the baseline model and our method on PACS. Different colors mean different classes in (a) and (b), and different domains in (c) and (d), respectively.}
			\label{Fig:5}
		\end{center}
	\end{figure}

\textbf{Further analysis on the progressive training scheme.} The results in Table \ref{tab:5} show the importance of introducing random noise in FSR module. Here, we present the results of applying FSR module when removing the extra noise to each individual block. Furthermore, we also conduct experiments for applying other general-purpose feature-based data manipulations over each block of the backbone with the proposed progressive training scheme. The results can be observed in Table \ref{tab:a1}, where ``w/o noise'' means FSR module without introducing extra noise, MixStyle~\cite{zhou2021domain} is simply mixing the style statistics on feature-level of the input images and pAdaIN~\cite{Nuriel2020PermutedAR} means randomly replacing the style statistics of the input images via a given probability. As shown in Table \ref{tab:5}, the proposed progressive training scheme can further improve the performance of all manipulations, but using our FSR can obtain the most improvement, which could owe to the learnable parameters and extra noise information in the proposed FSR module.
Therefore, the progressive training scheme can better promote the efficacy of the proposed FSR module when compared to other feature-level augmentation methods without learnable parameters and extra noise information.

\textbf{Sensitivity of the hyperparameters.} We conduct the experiment when training the network with various values of $\lambda$, \ie, the setting of 0, 0.4, 0.8, 1 and 2. Recall that $\lambda$ is used to trade off the two classification losses based on original features and augmented features in Eq.~(\ref{eq:6}). We report the different curve changes of all these settings in Fig. \ref{Fig:hp}. We first evaluate $\lambda$ on PACS, the results are reported in Fig. \ref{Fighp.sub.1}, where $\lambda = 0$ means the baseline model trained merely with original features. As shown in Fig. \ref{Fighp.sub.1}, when $\lambda$ is as small as 0.4, limited augmented features could be involved thus the diversity brought by FSR is inadequate, so that the performance is relatively poor. When $\lambda$ goes up from 0.4 to 1, the accuracy rises from 83.93\% to 85.95\%. However, when $\lambda$ is set to 2, the accuracy drops by 1.62\% (84.33\% vs. 85.95\%). This means that too large $\lambda$ could include excessive randomness to the model which make the model difficult to train, thus resulting in degradation in performance. We notice that it is similar to the value of $\lambda$ having the best overall accuracy for VLCS, as shown in Fig. \ref{Fighp.sub.2}. However, since the OfficeHome has 65 categories, the number of samples is not enough for full training, so more augmented features will bring much randomness to training, resulting in a decrease in accuracy, as shown in Fig. \ref{Fighp.sub.3}. The best result appears when $\lambda$ is 0.8, and continuing to increase $\lambda$ will make the result poor. Thus, we uniformly set $\lambda$ as 1 for PACS and VLCS, and as 0.8 for OfficeHome.

Moreover, we conduct the experiment by replacing the distribution of the mixing parameter $\alpha$ in Eq.~(\ref{eq:9}) and Eq.~(\ref{eq:10}) with the beta distribution, namely $\alpha \sim Beta(\lambda_{beta}, \lambda_{beta})$ where $\lambda_{beta} \in (0, \infty)$ is the parameter of Beta distribution as in \cite{zhou2021domain}. We report the results in \underline{\textbf{Table \ref{tab:beta}}}. As observed, the value of $\lambda$ has little effect on the final performance. This is because that, the proposed FSR including an encoder and a decoder aims to learn the new styles varying around the styles of source domains. The more diverse domain variations can be captured, the stronger generalization ability of the model. In this sense, our method focuses on the random breadth, which aims to the diversity of random styles, rather than the random depth, \ie, the specifically generated styles. Particularly, the random breadth has been considered by our method, \ie, adding the random noise into the original styles, therefore the distribution of random number $\alpha$ has little effect on the performance. Thus, we use uniform distribution in this paper.

\begin{table}[htbp]
	\renewcommand\arraystretch{1.3}
		\begin{center}
		\caption{Experimental results (\%) on using Beta distribution to mix the original style and random noise of PACS with ResNet-18. Each column indicates the name of unseen target domain.}
		\label{tab:beta}
		
			\footnotesize{
				\begin{tabular}{l|ccccc}
					\specialrule{0.8pt}{2pt}{2pt}
					Method & Art & Cartoon & Photo & Sketch & Avg\\
					\specialrule{0pt}{1pt}{1pt}
					\hline
					\specialrule{0pt}{1pt}{1pt}
					$\lambda_{beta} = 0.1$ & 83.62 & 80.61 & 95.63 & 81.23 & 85.27\\
					$\lambda_{beta} = 0.4$ & 83.67 & 80.95 & 95.45 & 81.64 & 85.43\\
					$\lambda_{beta} = 0.7$ & 84.13 & 80.89 & 95.65 & 81.23 & 85.53\\
					$\lambda_{beta} = 1.0$ & 84.49 & 81.15 & 96.13 & 82.01 & 85.95\\
					$\lambda_{beta} = 2.0$ & 83.57 & 80.18 & 95.87 & 81.04 & 85.17\\
					$\lambda_{beta} = 4.0$ & 83.79 & 81.02 & 96.17 & 81.02 & 85.50\\
					\specialrule{0.8pt}{2pt}{2pt}
			\end{tabular}}
		\end{center}
	\end{table}

\textbf{Visualization for feature representations.} We employ t-SNE \cite{Maaten2008VisualizingDU} to provide the visualization results of feature representations extracted by the learned feature extractor. We use ResNet-18 as backbone to conduct the experiment on PACS. Considering Photo as target domain and use the other three domains to train the model, the results of learned features of the baseline model and our proposed method are reported in Fig.~\ref{Fig:5}. 
	
It can be seen from the comparison between Fig. \ref{Fig5.sub.1} and Fig. \ref{Fig6.sub.1} where different colors denote different classes, the distance between different classes of our method is more obvious than the baseline model. This means that the features learned by our model have a clearer decision boundary, which is beneficial for the final classification. Furthermore, as we can see in Fig. \ref{Fig5.sub.2} where different colors represent different domains, the features learned by the baseline model have a relatively large gap between different domains. Specifically, Sketch is the farthest from the target domain Photo. On the contrary, in Fig. \ref{Fig6.sub.2}, there is no obvious domain gap based on the features learned by our method when compared with the baseline model. Thus, this reveals that our proposed method could indeed make the feature extractor pay more attention on domain-invariant information, which could make the model generalize well on unseen target domains.
	
\begin{figure}[!h]
  \centering
  \includegraphics[width=1\linewidth]{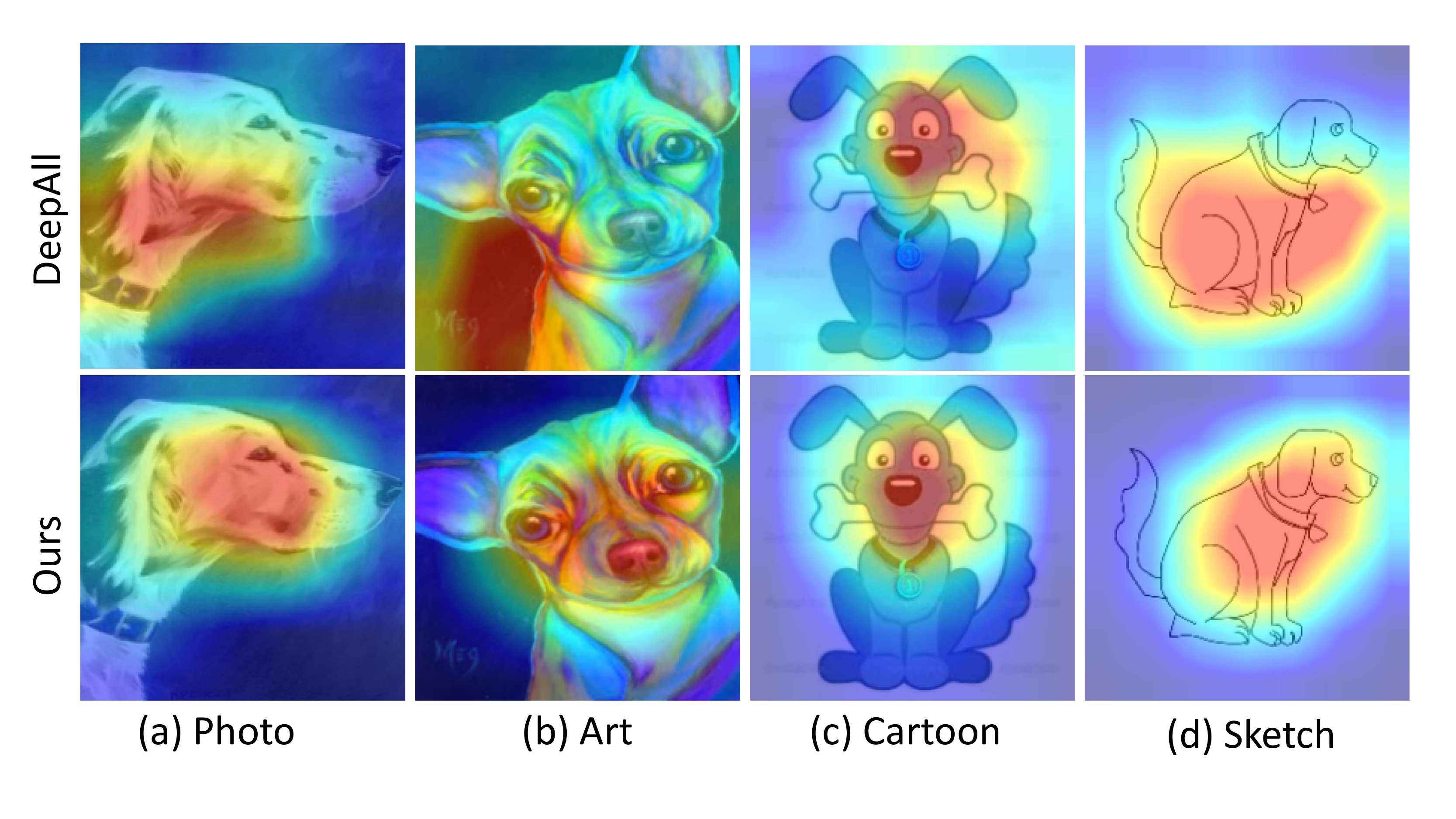}
  \vspace{-0.5cm}
  \caption{Visualization of different methods using GradCAM \cite{selvaraju2017grad} on the PACS dataset. Note that the redder the area indicates the higher attention. Top row is the baseline method (\ie, DeepAll), and bottom row is our method. The first three columns are source domains and the last is unseen target domain.}
  \label{data_vis}
\end{figure}

\textbf{Visualization for the class activation map.} We show the visualization results of activation maps using GradCAM \cite{selvaraju2017grad} on data level in Fig. \ref{data_vis}. As seen, our method stimulates the network to pay more attention on domain-invariant and discriminative features, \eg, \textit{our method tends to focus on some precise and comprehensive regions like ``head'' of dog for photo and cartoon, and ignore the background and body of the dog}. Therefore, as validated on visualization results on feature level, the results also reveal that our method could better extract domain-independent information, which could guide the model to generalize well in unseen target domains.

\begin{figure}[t]
    \centering
    \hspace{-0.5cm}
    \includegraphics[width=9cm,height=6cm]{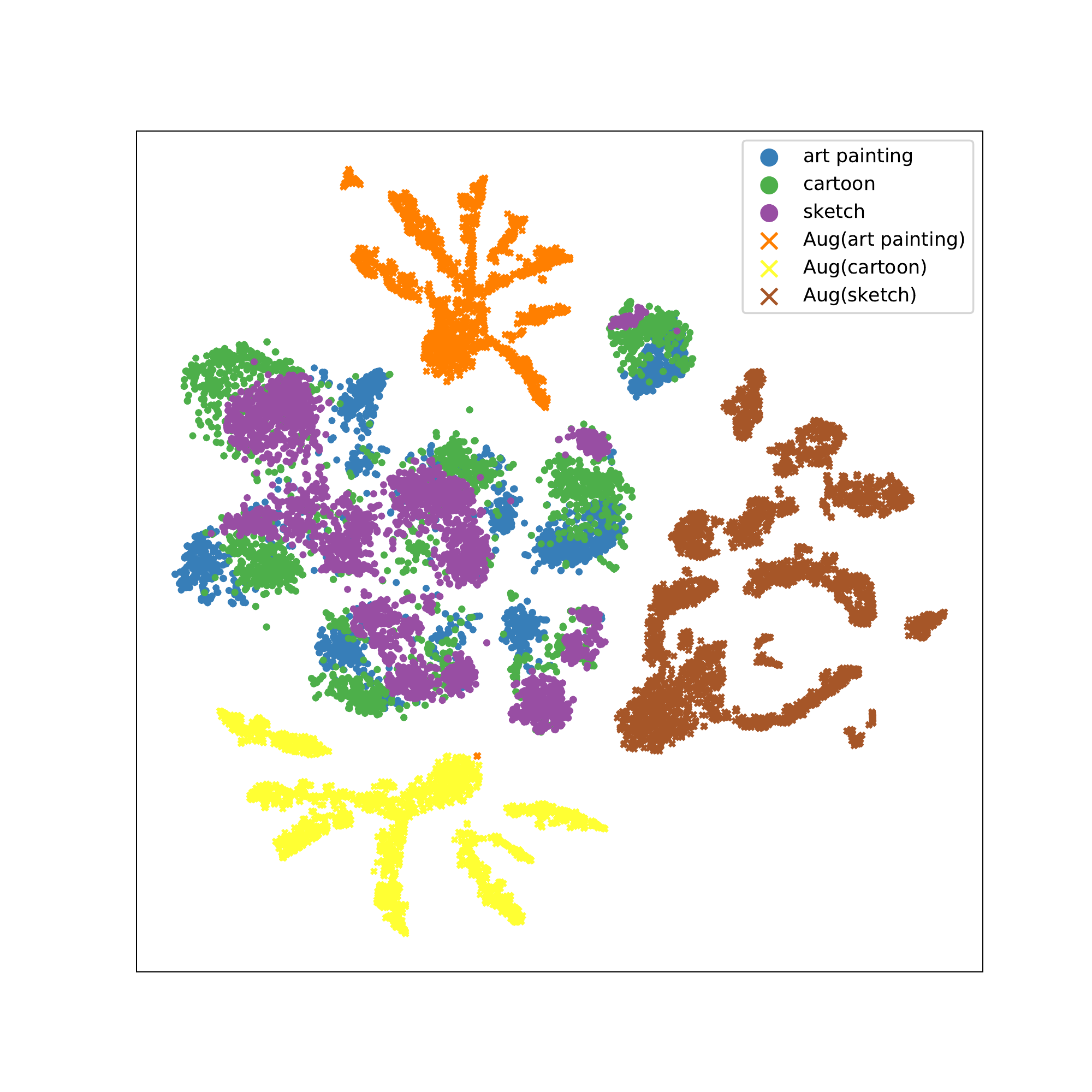}
    \vspace{-0.5cm}
    \caption{The visualization of the original features and the augmented features produced by our method on PACS when considering ``photo'' as the target domain. Different colors refer to different domains.}
    \label{domain-spec}
\end{figure}

\textbf{Visualization for the augmented features.} In this part, we provide the visualized results of the augmented features in our method. We can observe in Fig. \ref{domain-spec} that 1) the augmented features have different distributions from the original source domains while retaining the original category discriminability information and 2) the augmented features denoted by crosses are obviously separated by different domain labels, which can demonstrate the effectiveness of the proposed domain-specific FSR. Consequently, further training on these features improves the robustness of classification networks and achieves better capacity for domain generalization.

\textbf{Performance on source domains.} Table~\ref{tab:source} reports the results on the validation set of the source domains on PACS with the baseline model and our method, respectively. It can be observed that our method increases the results on the held-out validation set in all cases, thus this indicates our method can also increase the performance on the source domains while improving the generalization ability of the model on unseen domains.
	
\textbf{More results in medical imaging classification task.} We also evaluate our proposed method in the medical imaging classification task, \ie, skin lesion classification. We adopt six publicly available skin lesion datasets as different domains for experiments, including HAM10000 \cite{Tschandl2018TheHD}, Derm7pt (D7P) \cite{Kawahara2019SevenPointCA}, MSK \cite{Gutman2018SkinLA}, PH2 \cite{Mendona2013PH2A}, SONIC (SON) \cite{Gutman2018SkinLA}, and UDA \cite{Gutman2018SkinLA}. We follow the protocol in \cite{Yoon2019GeneralizableFL} by choosing seven-category subset from these datasets, including melanoma (mel), melanocytic nevus (nv), dermatofibroma (df), basal cell carcinoma (bcc), vascular lesion (vasc), benign keratosis (bkl), and actinic keratosis (akiec). And we use one dataset from D7P, MSK, PH2, SON and UDA as target domain and the remaining datasets together with HAM10000 as source domains. Each dataset is partitioned into 50\% training, 20\% validation, and 30\% testing. Besides, the value of hyper-parameter of $\lambda$ is set to 2.
We use the ResNet-18 model pretrained on ImageNet as the backbone. Other settings are same as described in Section~\ref{es1}. \\
\indent We compare our method with state-of-the-art domain generalization methods, including MixStyle~\cite{zhou2021domain} and pAdaIN~\cite{Nuriel2020PermutedAR}. The results are reported in Table \ref{tab:skin}, which is averaged by repeating three times. ``DeepAll'' refers to directly training model on the aggregation of source domains with classification loss. Note that all the methods leverage the advance of data augmentation to improve generalization ability of the model. As introduced in \cite{Zhang2020GeneralizingDL} and \cite{Li2020DomainGF}, the authors assume medical image domain variability can be conducted through linear transformation, \eg, blurriness, brightness, rotation and scaling changes. Li \etal \cite{Li2020DomainGF} resample the features in the latent space with linear dependency. However for MixStyle, it generates features by simply mixing the style statistics of original source domains. And for pAdaIN, it randomly switches the style statistics of the input images via a given probability. According to the experimental results, we observe that simply mixing or exchanging the style of source domains in the field of medical imaging will cause performance degradation when compared with the baseline model. Since the medical image domain variability is more compact than other image data, these augmentation methods may not be suitable for skin lesion classification task. However, our method adopts a learnable and adaptive augmentation way according to the provided domains and introduces random noise to bring more domain diversity to the model. In this way, it can not only retain consistent of some characteristics in medical images but also product more possible and comprehensive variants than linear transformation. As observed, our proposed algorithm can achieve better performance in a clear margin compared with the baseline model (81.31\% vs. 78.98\%).

\begin{table}[htbp]
	\renewcommand\arraystretch{1.3}
		\begin{center}
		\caption{Experimental results of accuracy (\%) in the skin lesion classification task using ResNet-18 as backbone. The title of each column indicates the name of unseen target domain, and the best result of each target domain is marked in bold.}
		\label{tab:skin}
	
			\footnotesize{
				\begin{tabular}{l|cccccc}	\specialrule{0.8pt}{2pt}{2pt}
				Method & D7P & MSK & PH2 & SON & UDA & Avg\\
				\specialrule{0pt}{1pt}{1pt}
				\hline
				\specialrule{0pt}{1pt}{1pt}
				DeepAll & 60.02 & 67.57 & 91.00 & 99.03 & 77.23 & 78.97\\	
				MixStyle \cite{zhou2021domain} & 60.14 & 73.24 & 84.00 & 99.01 & 74.53 & 78.18\\
				pAdaIN \cite{Nuriel2020PermutedAR} & 60.64 & 71.84 & 86.50 & 99.76 & 72.80 & 78.31\\
				\specialrule{0pt}{1pt}{1pt}
				\hline
				\specialrule{0pt}{1pt}{1pt}
				Our Method & \textbf{61.79} & 
				\textbf{73.83} & \textbf{92.55} & \textbf{99.81} & \textbf{78.55} & \textbf{81.31}\\
				\specialrule{0.8pt}{2pt}{2pt}
			\end{tabular}}
		\end{center}
	\end{table}

\section{Conclusion}
In this paper, we propose a feature-based style randomization module for the DG task. Instead of performing augmentation on image level, we convert the original style information into the random style using the encoder-decoder mode from the feature-level perspective, which could perform more random, abstract, and diverse transformations for the input image. Besides, we summarize the impact of our module in different locations of the network and propose a novel training strategy to make the network be better trained. By extensively evaluating the proposed method on various benchmark datasets, \ie, PACS, VLCS and OfficeHome, the experimental results demonstrate that our method outperforms other related state-of-the-art methods in terms of the generalization accuracy. 


%
%

\ifCLASSOPTIONcaptionsoff
  \newpage
\fi

\bibliographystyle{IEEEtran}
\bibliography{sigproc}

\end{document}